%% file: paper.tex
\documentclass[10pt,twocolumn,letterpaper]{article}

\usepackage{cvpr}              % To produce the CAMERA-READY version
\input{vcl-shortcuts.tex}

\graphicspath{{./fig/}}

\usepackage{enumitem}
\setlist{nosep}

\usepackage[pagebackref,breaklinks,colorlinks]{hyperref}

\usepackage[capitalize]{cleveref}
\crefname{section}{Sec.}{Secs.}
\Crefname{section}{Section}{Sections}
\Crefname{table}{Table}{Tables}
\crefname{table}{Tab.}{Tabs.}

\def\cvprPaperID{7258} % *** Enter the CVPR Paper ID here
\def\confName{CVPR}
\def\confYear{2023}

\begin{document}

%%%%%%%%% TITLE - PLEASE UPDATE
%
% Must include words:
% - Monocular
% - SDF
% - Reconstruction
% - Semantic
% - Continuous CRF

\title{Fast Monocular Scene Reconstruction with Global-Sparse Local-Dense Grids}

\author{
Wei Dong\thanks{CMU RI. Work done during the internship at NVIDIA.}\\
%{\small weidong@andrew.cmu.edu}
% For a paper whose authors are all at the same institution,
% omit the following lines up until the closing ``}''.
% Additional authors and addresses can be added with ``\and'',
% just like the second author.
% To save space, use either the email address or home page, not both
\and
Chris Choy\thanks{NVIDIA.}\\
\and
Charles Loop\footnotemark[2]\\
\and
Or Litany\footnotemark[2]\\
\and
Yuke Zhu\footnotemark[2]\\
\and
Anima Anandkumar\footnotemark[2]\\
}
\maketitle

%%%%%%%%% ABSTRACT
\begin{abstract}
Indoor scene reconstruction from monocular images has long been sought after by augmented reality and robotics developers. Recent advances in neural field representations and monocular priors have led to remarkable results in scene-level surface reconstructions. The reliance on Multilayer Perceptrons (MLP), however, significantly limits speed in training and rendering. 
In this work, we propose to directly use signed distance function (SDF) in sparse voxel block grids for fast and accurate scene reconstruction without MLPs. 
Our globally sparse and locally dense data structure exploits surfaces' spatial sparsity, enables cache-friendly queries, and allows direct extensions to multi-modal data such as color and semantic labels.
To apply this representation to monocular scene reconstruction, we develop a scale calibration algorithm for fast geometric initialization from monocular depth priors. We apply differentiable volume rendering from this initialization to refine details with fast convergence. 
We also introduce efficient high-dimensional Continuous Random Fields (CRFs) to further exploit the semantic-geometry consistency between scene objects.
Experiments show that our approach is 10$\times$ faster in training and 100$\times$ faster in rendering while achieving comparable accuracy to state-of-the-art neural implicit methods.
\end{abstract}

\section{Introduction}
\input{tex/intro.tex}

\section{Related Work}
\input{tex/related.tex}

\section{Method}
\input{tex/method_overview.tex}
\input{tex/method_calibration.tex}

\input{tex/method_fusion.tex}
\input{tex/method_refinement.tex}

\section{Experiments}
\input{tex/experiment.tex}

\section{Conclusion}
\input{tex/conclusion.tex}

%%%%%%%%% REFERENCES
{\small
\bibliographystyle{ieee_fullname}
\bibliography{paper}
}

\newpage
\input{tex/supplement_sec}

\end{document}

%% file: vcl-shortcuts.tex
\usepackage{epsfig}
\usepackage{graphicx}
\usepackage{float}
\usepackage{wrapfig}
\usepackage{amsmath,amssymb,amsthm}
\usepackage{algorithm,algorithmicx,algpseudocode}
\usepackage{bm,xspace}
\usepackage{comment}
\usepackage{verbatim}
\usepackage{multirow}
\usepackage{balance}
\usepackage{url}
\usepackage{booktabs}
\usepackage{etoolbox,siunitx}
\usepackage{calc}
\usepackage{pifont,hologo}
\usepackage[usenames, dvipsnames]{color}
\usepackage{nicefrac}
\usepackage{listings}
\usepackage{xcolor}
\usepackage{framed}
\usepackage{enumitem}
\usepackage{colortbl}
\usepackage{resizegather}

\newcommand\scalemath[2]{\scalebox{#1}{\mbox{\ensuremath{\displaystyle #2}}}}

\newcommand{\one}[1]{\cellcolor{yellow!75}\textbf{#1}}
\newcommand{\two}[1]{\cellcolor{yellow!25}{#1}}

\definecolor{codegreen}{rgb}{0,0.6,0}
\definecolor{codegray}{rgb}{0.5,0.5,0.5}
\definecolor{codepurple}{rgb}{0.58,0,0.82}
\definecolor{backcolor}{rgb}{0.89,0.99,0.97}
\lstdefinestyle{mystyle}{
    backgroundcolor=\color{backcolor},
    commentstyle=\color{codegreen},
    keywordstyle=\color{magenta},
    numberstyle=\tiny\color{codegray},
    basicstyle=\ttfamily\footnotesize,
    keepspaces=true,
    numbers=left,
    numbersep=3pt,
    belowcaptionskip=5em,
    belowskip=3em,
    numbers=left,
    xleftmargin=0.8em,
    columns=fullflexible,
}
\usepackage[super]{nth}
\usepackage{nicefrac}
\robustify\bfseries

\def\cc{\mathbf{c}}

\def\ff{\mathbf{f}}

\def\pp{\mathbf{p}}

\def\rr{\mathbf{r}}
\def\sss{\mathbf{s}}
\def\tt{\mathbf{t}}

\def\vv{\mathbf{v}}

\def\xx{\mathbf{x}}

\def\RR{\mathbf{R}}

\def\XX{\mathbf{X}}

\def\dD{\mathcal{D}}

\def\iI{\mathcal{I}}

\def\lL{\mathcal{L}}

\def\nN{\mathcal{N}}

\def\sS{\mathcal{S}}

\DeclareMathOperator*{\argmin}{arg\,min}
\DeclareMathOperator*{\mean}{mean}

\DeclareMathSymbol{@}{\mathord}{letters}{"3B}

\def\latex/{\LaTeX}
\def\bibtex/{\hologo{BibTeX}}

\def\eg{\textit{e.g.}}

%% file: tex/intro.tex
% Main paragraph idea at the top and the corresponding content

% - Indoor / large scale reconstruction is an important problem used in XXX.
Reconstructing indoor spaces into 3D representations is a key requirement for many real-world applications, including robot navigation, immersive virtual/augmented reality experiences, and architectural design. Particularly useful is reconstruction from monocular cameras which are the most prevalent and accessible to causal users. While much research has been devoted to this task, several challenges remain. 
\begin{figure}[ht]
    \centering
    \includegraphics[width=\columnwidth]{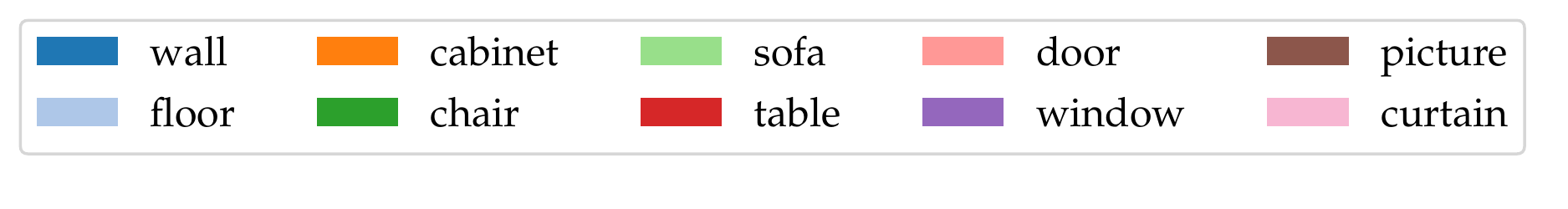}
    \begin{tabular}{@{}c@{}c@{}}
    \includegraphics[width=.5\columnwidth]{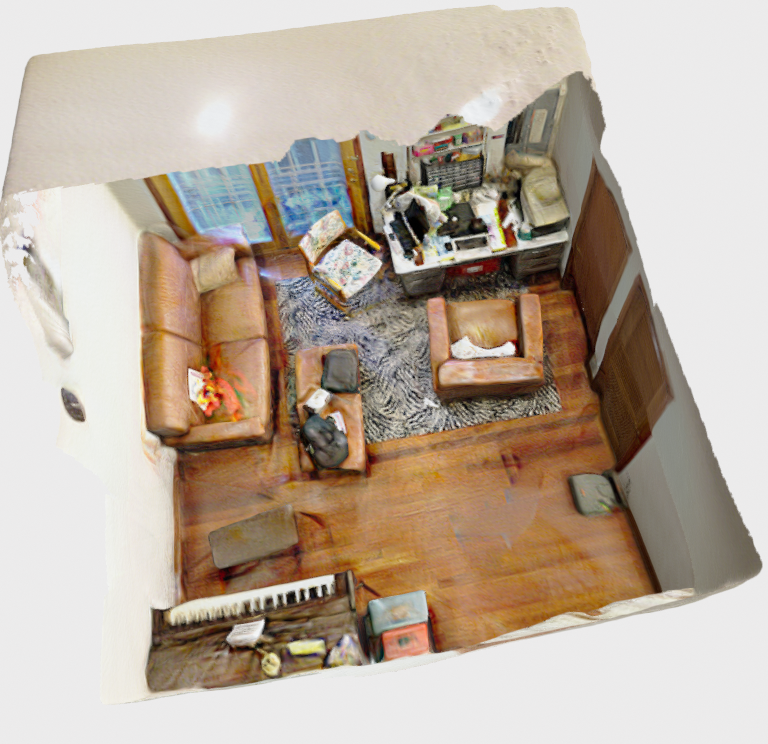} &
    \includegraphics[width=.5\columnwidth]{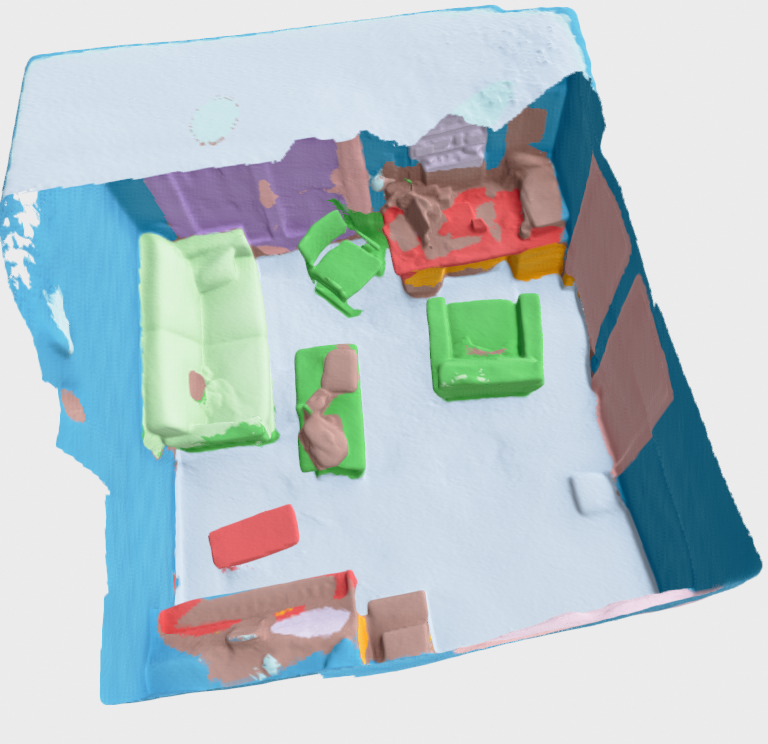}    
    \end{tabular}
    \caption{Color and semantic scene reconstruction from our system with monocular images and learned monocular priors.
    }
    \label{fig:semantic}
\end{figure}

\begin{figure*}[t]
    \centering
    \footnotesize
    \begin{tabular}{c@{}c@{}c@{}c}
    \includegraphics[width=.24\textwidth]{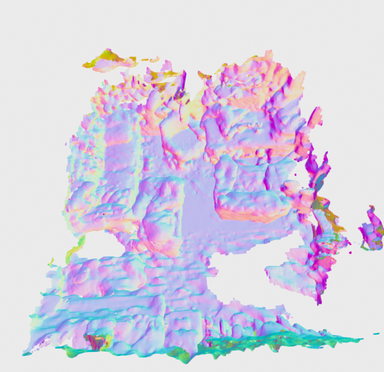} &
    \includegraphics[width=.24\textwidth]{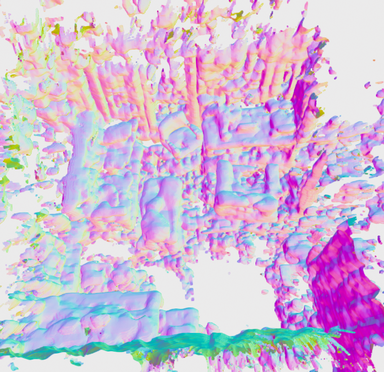} &
    \includegraphics[width=.24\textwidth]{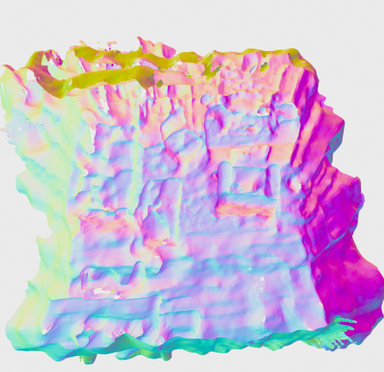} &
    \includegraphics[width=.24\textwidth]{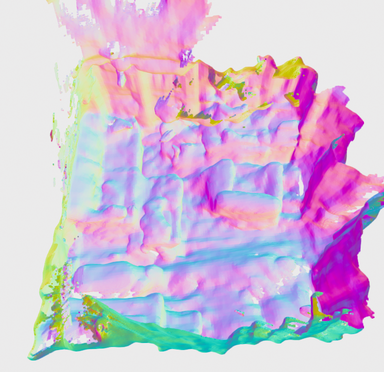} \\
    (a) COLMAP~\cite{schonberger2016structure} & (b) NeRF~\cite{mildenhall2021nerf} & (c) VolSDF~\cite{yariv2021volume} & (d) NeuS~\cite{wang2021neus} \\
    \includegraphics[width=.24\textwidth]{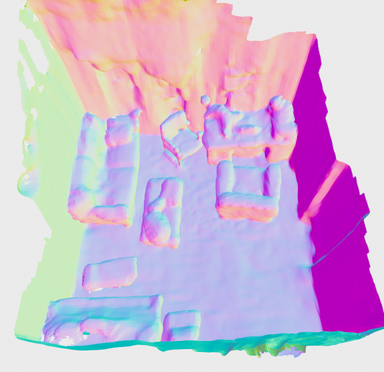} &
    \includegraphics[width=.24\textwidth]{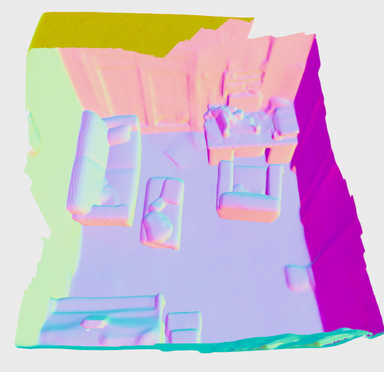} &
    \includegraphics[width=.24\textwidth]{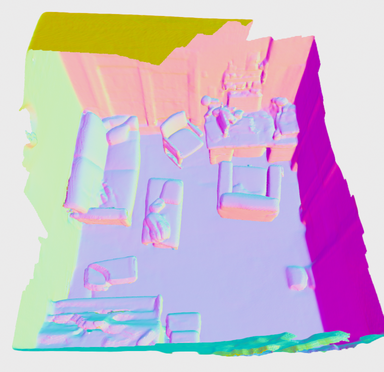} &
    \includegraphics[width=.24\textwidth]{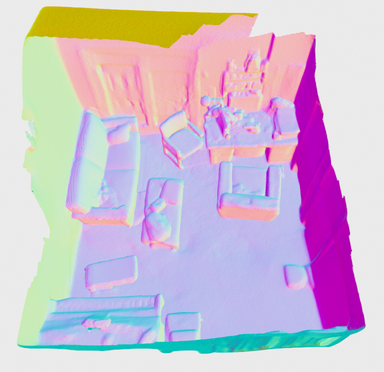} \\
    (e) ManhattanSDF~\cite{guo2022neural} & (f) MonoSDF-MLP~\cite{yu2022monosdf} & (g) MonoSDF-Grid~\cite{yu2022monosdf} & (h) Ours
    \end{tabular}
    \caption{Qualitative reconstruction comparison on ScanNet~\cite{dai2017scannet}. While being 10$\times$ faster in training, we achieve similar reconstruction results to state-of-the-art MonoSDF~\cite{yu2022monosdf}, with fine details (see Fig.~\ref{fig:details-0050}).}
    \label{fig:qualitative-0050}
\end{figure*}

Conventional monocular reconstruction from multi-view RGB images uses patch matching~\cite{schonberger2016structure}, which takes hours to reconstruct even a relatively small scene.
Several 3D reconstruction methods~\cite{yao2018mvsnet, sun2021neuralrecon} have demonstrated fast reconstruction by applying 3D convolutional neural networks to feature volumes, but they have limited resolution and struggle to generalize to larger scenes.

Recently, unified neural radiance fields~\cite{mildenhall2021nerf} and neural implicit representations were developed for the purpose of accurate surface reconstruction from images~\cite{oechsle2021unisurf,wang2021neus,yariv2021volume}. 
While this was successfully demonstrated on single objects, the weak photometric constraint leads to poor reconstruction and slow convergence for large-scale scenes. Guo \etal~\cite{guo2022neural} and Yu \etal~\cite{yu2022monosdf} improved the quality and convergence speed of neural field reconstruction on large-scale scenes by incorporating learned geometrical cues like depth and normal estimation~\cite{ranftl2021vision, eftekhar2021omnidata}, however, training and evaluation remain inefficient. This is primarily because these approaches rely on MLPs and feature grids~\cite{muller2022instant} that encode the entire scene rather than concentrating around surfaces.

In contrast to MLPs, an explicit SDF voxel grid can be adaptively allocated around surfaces, and allows fast query and sampling. 
However, an efficient implementation of differentiable SDF voxel grids without MLPs is missing. 
Fridovich-Keil and Yu \etal~\cite{fridovich2022plenoxels} used an explicit density and color grid, but is limited to rendering small objects. 
Muller \etal~\cite{muller2022instant} developed a feature grid with spatial hashing for fast neural rendering, but its backbone hash map is not collision-free, causing inevitable slow random access and inaccurate indexing at large scales.
Dong \etal~\cite{dong2021ash} proposed a collision-free spatially hashed grid following Niessner \etal~\cite{niessner2013real}, but lacks support for differentiable rendering. 
Several practical challenges hinder the implementation of an efficient differentiable data structure: 1. a collision-free spatial hash map on GPU that supports one-to-one indexing from positions to voxels; 2. differentiable trilinear interpolations between spatially hashed voxels; 3. parallel ray marching and uniform sampling from a spatial hash map.

\textbf{Our approach:} we address such challenges using a differentiable \emph{globally sparse} and \emph{locally dense} voxel grid.
We transform a collision-free GPU hash map~\cite{stotko2019slamcast} to a differentiable tensor indexer~\cite{paszke2017automatic}.
This generates a one-to-one map between positions and \emph{globally sparse} voxel blocks around approximate surfaces, and enables skipping empty space for efficient ray marching and uniform sampling.
We further manage \emph{locally dense} voxel arrays within sparse voxel blocks for GPU cache-friendly contiguous data query via trilinear interpolation.
As a result, using explicit SDF grids leads to fast SDF gradient computation in a single forward pass, which can further accelerate differentiable rendering.

% Depth scaler
This new data structure presents a new challenge --- we can only optimize grid parameters if they are allocated around surfaces. To resolve this, we make use of off-the-shelf monocular depth priors~\cite{ranftl2021vision,eftekhar2021omnidata} and design a novel initialization scheme
with global structure-from-motion (SfM) constraints to calibrate these unscaled predicted depths. It results in a consistent geometric initialization via volumetric fusion  ready to be refined through differentiable volume rendering.

We additionally incorporate semantic monocular priors~\cite{li2022language} to provide cues for geometric refinement in 3D. For instance, we use colors and semantics to guide the sharpening of normals around object boundaries, which in turn improves the quality of colors and semantics. 
We enforce these intuitive notions through our novel continuous Conditional Random Field (CRF). We use Monte Carlo samples on the SDF zero-crossings to create continuous CRF nodes and define pairwise energy functions to enforce local consistency of colors, normals, and semantics. 
Importantly, we define similarity in a \emph{high dimensional space} that consists of coordinates, colors, normals, and semantics, to reject spatially close samples with contrasting properties.
To make inference tractable, we follow Krahenbuhl \etal~\cite{krahenbuhl2011efficient} and use variational inference, leading to a series of convolutions in a high-dimensional space. We implement an efficient permutohedral lattice convolution~\cite{adams2010fast} using the collision-free GPU hashmap to power the continuous CRF inference.

The final output of our system is a scene reconstruction with geometry, colors, and semantic labels, as shown in Fig.~\ref{fig:semantic}. Experiments show that our method is $10\times$ faster in training, $100\times$ faster in inference, and has comparable accuracy measured by F-scores against state-of-the-art implicit reconstruction systems~\cite{guo2022neural,yu2022monosdf}. 
In summary, we propose a fast scene reconstruction system for monocular images. Our contributions include:
\begin{itemize}[leftmargin=*]
  \item A globally sparse locally dense differentiable volumetric data structure that exploits surface spatial sparsity %with explicit SDF storage;
  without an MLP;
  \item A scale calibration algorithm that produces consistent geometric initialization from unscaled monocular depths;
  \item A fast monocular scene reconstruction system equipped with volume rendering and high dimensional continuous CRFs optimization.
%\orl{should we split the CRF to a separate contribution?}
\end{itemize}

%% file: tex/related.tex
\noindent \textbf{Surface reconstruction from 3D data.}
Surface reconstruction has been well-studied from 3D scans.
The general idea is to represent the space as an implicit signed distance function, and recover surfaces at zero-crossings with Marching Cubes~\cite{lorensen1987marching}.
Classical works~\cite{curless1996volumetric, newcombe2011kinectfusion, niessner2013real, dai2017bundlefusion, dong2021ash} quantize the 3D space into voxels, and integrate frame-wise SDF observations into voxels.
Instead of direct voxels, recent neural representations~\cite{sucar2021imap, zhu2022nice, takikawa2021nglod} use either a pure MLP or a feature grid to reconstruct smoother surfaces.
These approaches are often fast and accurate, but heavily depends on high-quality depth input from sensors.

\noindent \textbf{Surface reconstruction from RGB.}
A variety of classical and learning-based methods~\cite{newcombe2011dtam, yao2018mvsnet, sun2021neuralrecon, zhang2022nerfusion} have been developed to achieve high quality multi-view depth reconstruction from monocular images. These techniques usually construct a cost volume between a reference frame and its neighbor frames, and maximize the appearance consistency. A global volume can be optionally grown from the local volumes~\cite{murez2020atlas, sun2021neuralrecon, zhang2022nerfusion} for scene reconstruction. While these approaches succeed on various benchmarks, they rely on fine view point selection, and the performance may be significantly reduced when the view points and surfaces are sparse in space. Training on a large datasets is also required.

Recent advances in neural rendering~\cite{tulsiani2017multi, mildenhall2021nerf, barron2021mip} and their predecessors have defined the surface geometry by a density function predicted by an MLP in 3D space. They seek to minimize ray consistency with the rendering loss using test-time optimization. While being able to achieve high rendering quality, due to the ambiguity in the underlying density representation, accurate surfaces are hard to recover. In view of this, implicit SDF representations~\cite{yariv2020multiview, yariv2021volume, wang2021neus, sun2022neural} are used to replace density, where surfaces are better-defined at zero-crossings. To enable large-scale indoor scene reconstruction, ManhattanSDF~\cite{guo2022neural} and MonoSDF~\cite{yu2022monosdf} incorporate monocular geometric priors and achieve state-of-the-art results.
These approaches initialize the scene with a sphere~\cite{atzmon2020sal}, and gradually refine the details. As a result, the training time can be long, varying from hours to half a day.

\noindent \textbf{Monocular priors in surface reconstruction.}
Priors from monocular images have been used to enhance reconstruction and neural rendering from images, by providing reasonable initialization and better constrained sampling space.
A light weight prior is the structure-from-motion (SfM) supervision~\cite{schonberger2016structure}, where poses and sparse point clouds are reconstructed to provide the geometry. Similarly, dense monocular priors including depths~\cite{li2019learning, ranftl2021vision, ranftl2020towards}, normals~\cite{eftekhar2021omnidata}, and semantic segmentations~\cite{li2022language}.
Existing approaches either use only SfM~\cite{kangle2021dsnerf}, or enhance dense priors by SfM via finetuning depth prediction networks~\cite{wei2021nerfingmvs} or depth completion networks~\cite{roessle2022dense} for density-based neural rendering. Similarly, monocular priors are used to enhance SDF-based neural reconstruction~\cite{yu2022monosdf, guo2022neural} with remarkable performance. While emphasizing the guidance in sampling, these approaches usually stick to MLPs or dense voxel grids, without being able to fully exploit the sparsity of the surface distribution.

\noindent \textbf{Sparse spatial representations.}
Sparse spatial representations have been well-studied for 3D data, especially point clouds~\cite{choy20194d, niessner2013real, dong2021ash}. Often used data structures are hash maps, Octrees~\cite{meagher1982geometric}, or a combination~\cite{museth2013openvdb}.
These data structures have been adapted to neural 3D reconstructions and rendering to exploit spatial sparsity, but they either depend on high quality 3D input~\cite{takikawa2021nglod}, or focus on object-centered reconstruction~\cite{muller2022instant, clark2022volumetric, fridovich2022plenoxels}. Their usage to scene reconstruction from monocular images is yet to be explored.

%% file: tex/method_overview.tex
\begin{figure}[t]
    \centering
    \includegraphics[width=.95\columnwidth]{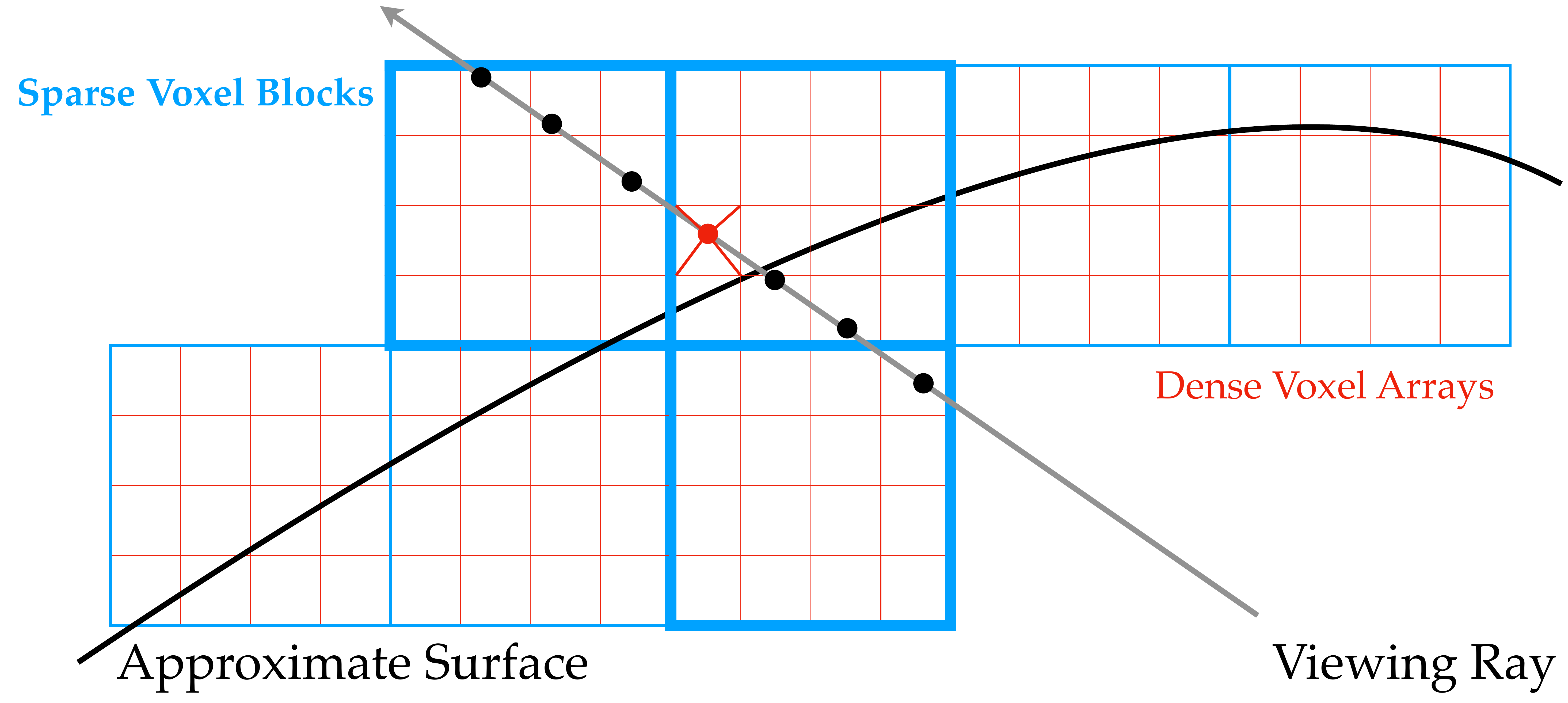}
    \vspace{1mm}
    \caption{Illustration of the sparse-dense data structure. The large voxel blocks (in blue) are allocated only around approximate surfaces, and indexed by a collision-free hash map. The voxel arrays (in red) further divide the space to provide high-resolution details. Ray marching skips empty space and only activates hit blocks (in bold blue). Trilinear interpolation of neighbor voxel properties allows sampling at continuous locations. }
    \label{fig:voxelblock}
\end{figure}

\begin{figure*}[t]
  \includegraphics[width=\textwidth]{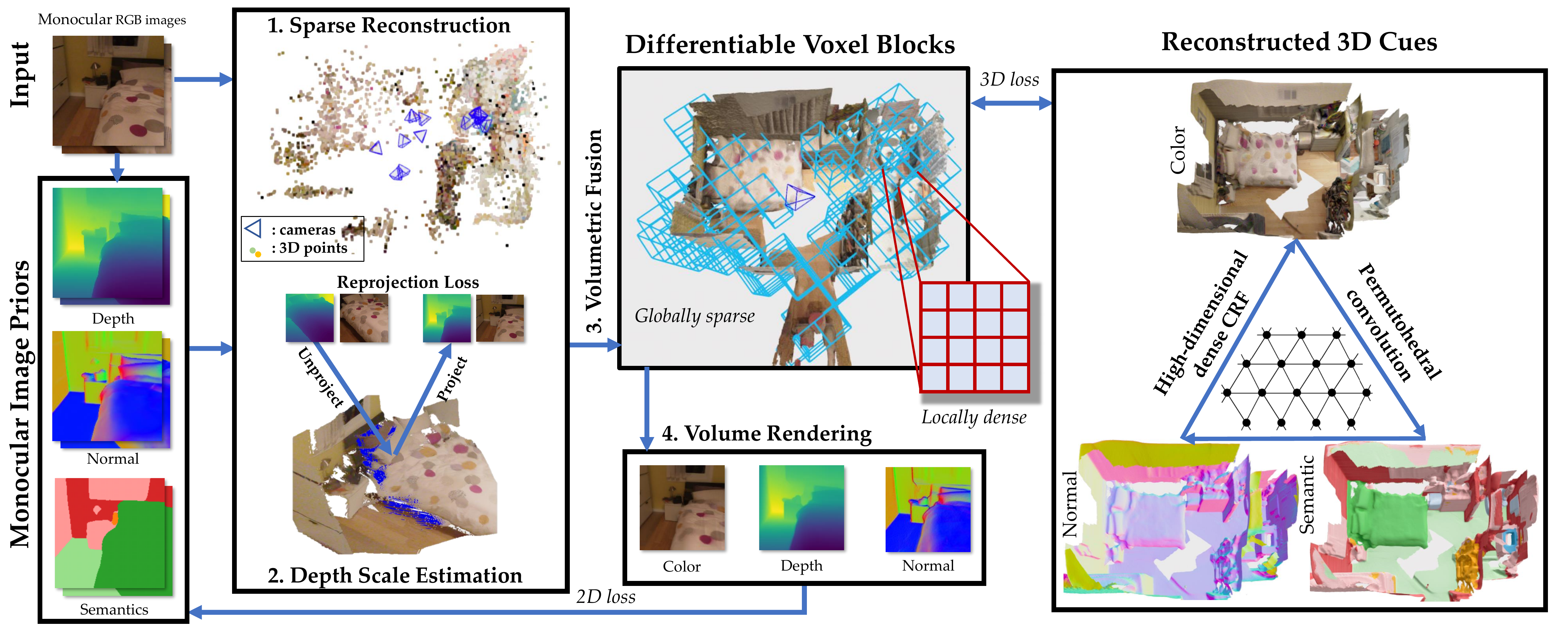}
  \caption{
  Illustration of our pipeline. We first use structure-from-motion (SfM) to obtain sparse feature-based reconstruction. With the sparse point cloud and covisibility information from SfM, we optimize the scale of predicted monocular depth images (\S \ref{subsec:scaler}), and perform volumetric fusion to construct a globally sparse locally dense voxel grid (\S \ref{subsec:fusion}). After initialization, we perform differentiable volume rendering to refine the details (\S \ref{subsubsect:rendering}), and apply high dimensional continuous CRFs to finetune normals, colors, and labels (\S \ref{subsubsec:crf}).}\label{fig:pipeline}
\end{figure*}
\subsection{Overview}

The input to our method is a sequence of monocular images $\{\iI_i\}$. Prior to reconstruction, similar to previous works~\cite{yu2022monosdf,guo2022neural}, we generate per-image monocular priors including unscaled depth $\{\dD_i\}$ and normal $\{\nN_i\}$ predicted by Omnidata~\cite{eftekhar2021omnidata}, and semantic logits $\{\sS_i\}$ from LSeg~\cite{li2022language}.
Afterwards, the system runs in three major stages.
\begin{itemize}[noitemsep,leftmargin=*]
\item Sparse SfM reconstruction~\cite{schonberger2016structure} and initial depth scale optimization;
\item Direct volumetric fusion for sparse voxel allocation and geometric initialization;
\item Differentiable volume rendering and dense CRF smoothing for detail refinement.
\end{itemize}
Fig.~\ref{fig:pipeline} shows the pipeline of our framework. We will describe these stages in detail after introducing our core data structure.

\subsection{Sparse-Dense Data Structure}
In order to facilitate multi-view sensor fusion, SDF are approximated by truncated SDF (TSDF) that maintain averaged signed distances to surface in a narrow  band close to the surface~\cite{curless1996volumetric, newcombe2011kinectfusion, niessner2013real}. 
We take advantage of this property and develop a globally sparse locally dense data structure.
Global sparsity is attained through allocating voxels only around approximate surfaces, which we index using a collision-free hash map. Within these sparse voxels we allocate 
cache-friendly small dense arrays that allow fast indexing and neighbor search storing SDF, color, and optionally labels. The data structure is visualized in Fig.~\ref{fig:voxelblock}.

While similar structures have been used for RGB-D data that focus on forward fusion~\cite{niessner2013real,dong2021ash}, our implementation supports both forward fusion via hierarchical indexing, and auto-differentiable backward optimization through trilinear interpolation, allowing refinement through volume rendering. In addition, SDF gradients can be explicitly computed along with SDF queries in the same pass, allowing efficient regularization during training.

Our data structure is akin to any neural networks that maps a coordinate $\xx \in \mathbb{R}^3$ to a property~\cite{xie2022neural}, thus in the following sections, we refer to it as a function $f$. We use $f_{\theta_d}, f_{\theta_c}, f_{\theta_s}$ to denote functions that query SDF, color, and semantic labels from the data structure, respectively, where $\theta_d, \theta_c$, and $\theta_s$ are properties directly stored at voxels.

%% file: tex/method_calibration.tex
\subsection{Depth Scale Optimization}\label{subsec:scaler}
Our sparse-dense structure requires approximate surface initialization at the allocation stage, hence we resort to popular monocular geometry priors~\cite{eftekhar2021omnidata} also used in recent works~\cite{yu2022monosdf}.
Despite the considerable recent improvement of monocular depth prediction, there are still several known issues in applications: each image's depth prediction is scale-ambiguous, often with strong distortions.
However, to construct an initial structure we require a consistent scale. %Without scale calibration, a consistent initial model cannot be constructed.

To resolve this, we define a scale function $\phi_i$ per monocular depth image $\dD_i$ to optimize scales and correct distortions. $\phi_i$ is represented by a 2D
grid, where each grid point stores a learnable scale. A pixel's scale $\phi_i(\pp)$ can be obtained through bilinear interpolating its neighbor grid point scales. 
We optimize $\{\phi_i\}$ to achieve consistent depth across frames
\begin{align}
  \min_{\{\phi_i\}} \sum_{{i, j} \in \Omega} h(\phi_i, \phi_j) + \lambda \sum_i g(\phi_i),
\end{align}
where $h$ and $g$ impose mutual and binary constraints, and $\Omega$ is the set of covisible image pairs.

Previous approaches~\cite{kopf2021robust} use fine-grained pixel-wise correspondences to construct $h$ via pairwise dense optical flow, and introduce a regularizer $g$ per frame.
This setup is, however, computationally intensive and opposes our initial motivation of developing an efficient system.
Instead, we resort to supervision from SfM's~\cite{schonberger2016structure} sparse reconstruction.
It estimates camera poses $\{R_i, t_i\}$, produces sparse reconstruction with 3D points $\{\xx_k\}$ and their associated 2D projections $\pp_{\xx_k \to i}$ at frame $\{\iI_i, \dD_i\}$, and provides the covisible frame pair set $\Omega$. With such, we can define the unary constraint $g$ via a reprojection loss
\begin{align}
&  g(\phi_i) = \sum_{\xx_k} \lVert d_{\xx_k \to i} - \dD_i(\pp_{\xx_k \to i}) \phi_i(\pp_{\xx_k \to i})\rVert^2, \\
&  d_{\xx_k \to i} \cdot \begin{bmatrix} \pp_{\xx_k \to i} & 1\end{bmatrix}^\top \triangleq \Pi \left(\RR_{i}^\top (\xx_k - \tt_{i})\right), \label{eq:reproj_3d}
\end{align}
where $\Pi$ is the pinhole projection.
Similarly, we define binary constraints by minimizing reprojection errors across covisible frames:
\begin{align}
  h(\phi_i, \phi_j) &= \sum_{\pp \in \dD_i} \lVert d_{i\to j} - \dD_j(\pp_{i\to j}) \phi_j(\pp_{i\to j}) \rVert^2 \nonumber \\ &
  + \lVert \iI_i(\pp) - \iI_j(\pp_{i\to j}) \lVert^2,\\
  \xx &= \Pi^{-1}\bigg(\pp, \dD_i(\pp) \phi_i (\pp) \bigg), \\
  d_{i\to j} \cdot \begin{bmatrix} \pp_{i \to j} & 1\end{bmatrix}^\top &\triangleq \Pi (\RR_{i,j} \xx + \tt_{i,j}),
\end{align}
where $\Pi^{-1}$ unprojects a pixel $\pp$ in frame $i$ from deformed depth to a 3D point $\xx$, and $\{\RR_{i,j}, \tt_{i,j}\}$ are relative poses. This loss enforces local consistency between visually adjacent frames.

We use a $24 \times 32$ 2D grid per image, $\lambda=10^{-3}$, and optimize $\{\phi_i\}$ via RMSprop with a learning rate of $10^{-2}$ for $500$ steps.

%% file: tex/method_fusion.tex
\begin{figure}[t]
    \centering
    \begin{tabular}{@{}c@{}c@{}}
    \includegraphics[width=.49\linewidth]{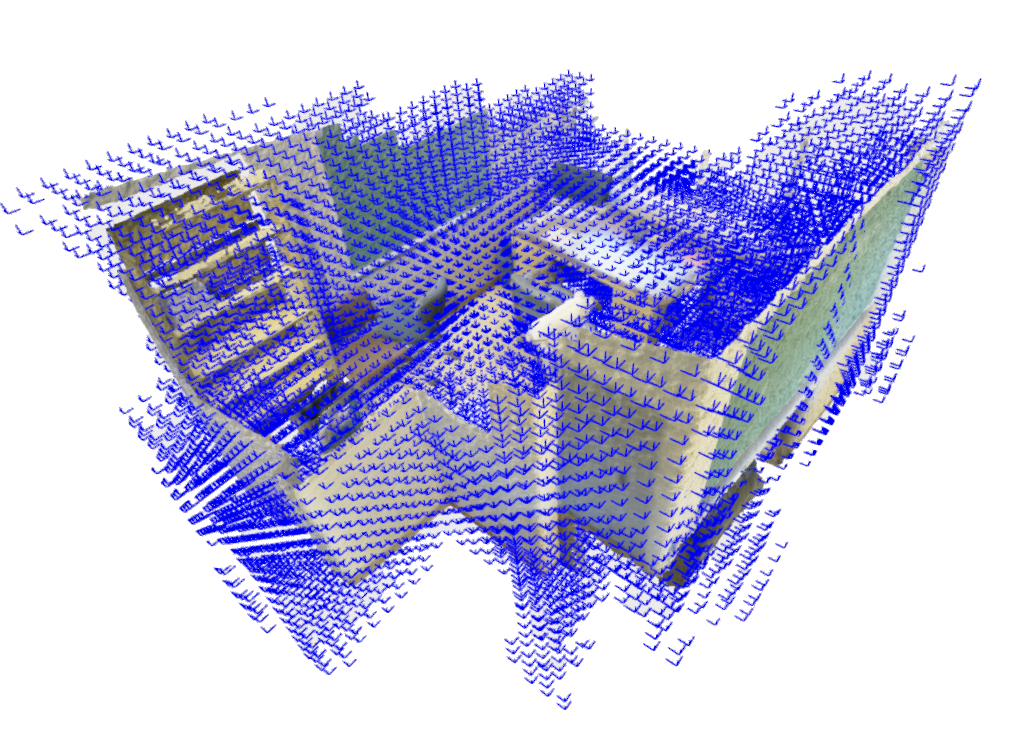} &
    \includegraphics[width=.49\linewidth]{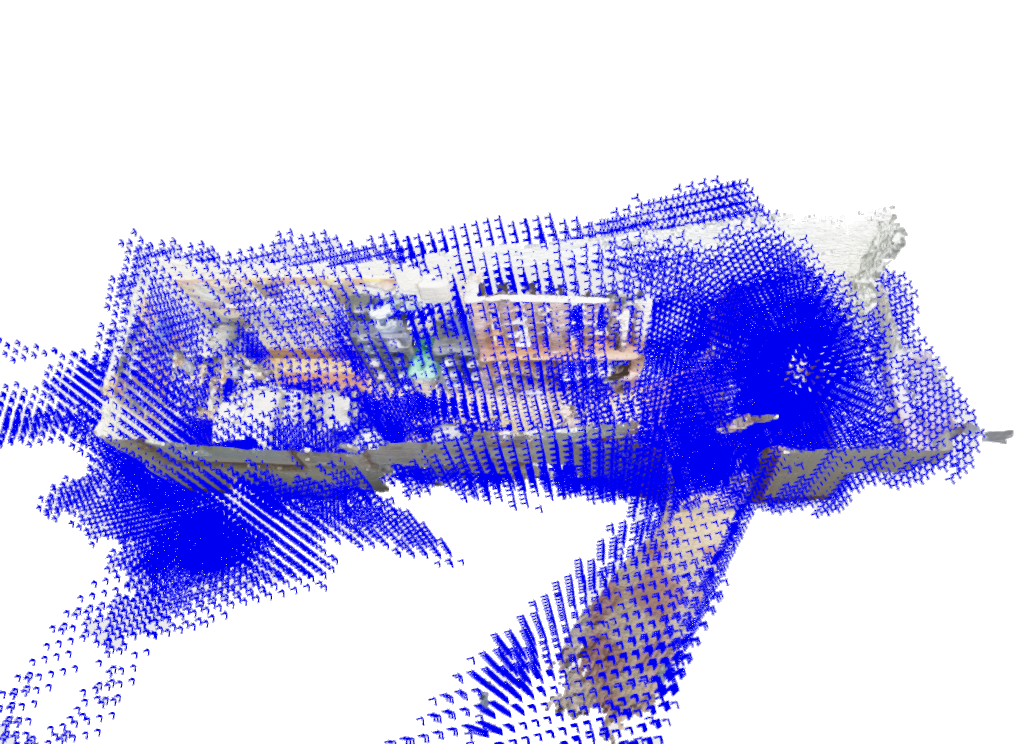}
    \end{tabular}
    \caption{Sparse voxel grid (blue) allocation around 3D points from unprojection. 
    The grids are adaptive to scenes with different overall surface shapes. Ground truth surface mesh are visualized for illustration.}
    \label{fig:sparse-grid}
\end{figure}

\subsection{Direct Fusion on Sparse Grid}\label{subsec:fusion}

\subsubsection{Allocation}
Similar to aforementioned works for online reconstruction~\cite{niessner2013real, dong2021ash}, the sparse blocks are allocated by the union of voxels containing the unprojected points,
\begin{align}
    \XX &= \cup_i \XX_i, \XX_i = \cup_p \bigg\{\textrm{Dilate} \left(\mathrm{Voxel}(\pp)\right) \bigg\}, \\
    \mathrm{Voxel}(\pp) &= \bigg\lfloor \frac{\RR_i \Pi^{-1} (\pp, \dD_i(\pp)\phi_i(\pp) ) + \tt_i}{L} \bigg\rfloor ,
\end{align}
where $L$ is the sparse voxel block size, and the dilate 
operation grows a voxel block to include its neighbors to tolerate more uncertainty from depth prediction.
A dynamic collision-free hash map~\cite{dong2021ash} is used to efficiently aggregate the allocated voxel blocks. The dense voxel arrays are correspondingly allocated underlying the sparse voxel blocks.

Fig.~\ref{fig:sparse-grid} shows the surface-adaptive allocation. In contrast to popular sparse grids in a fixed bounding box used by neural rendering~\cite{fridovich2022plenoxels, muller2022instant}, this allocation strategy is more flexible to various shapes of rooms.
\begin{figure}[t]
    \centering
    \begin{tabular}{@{}cc@{}}
    \includegraphics[width=.48\linewidth]{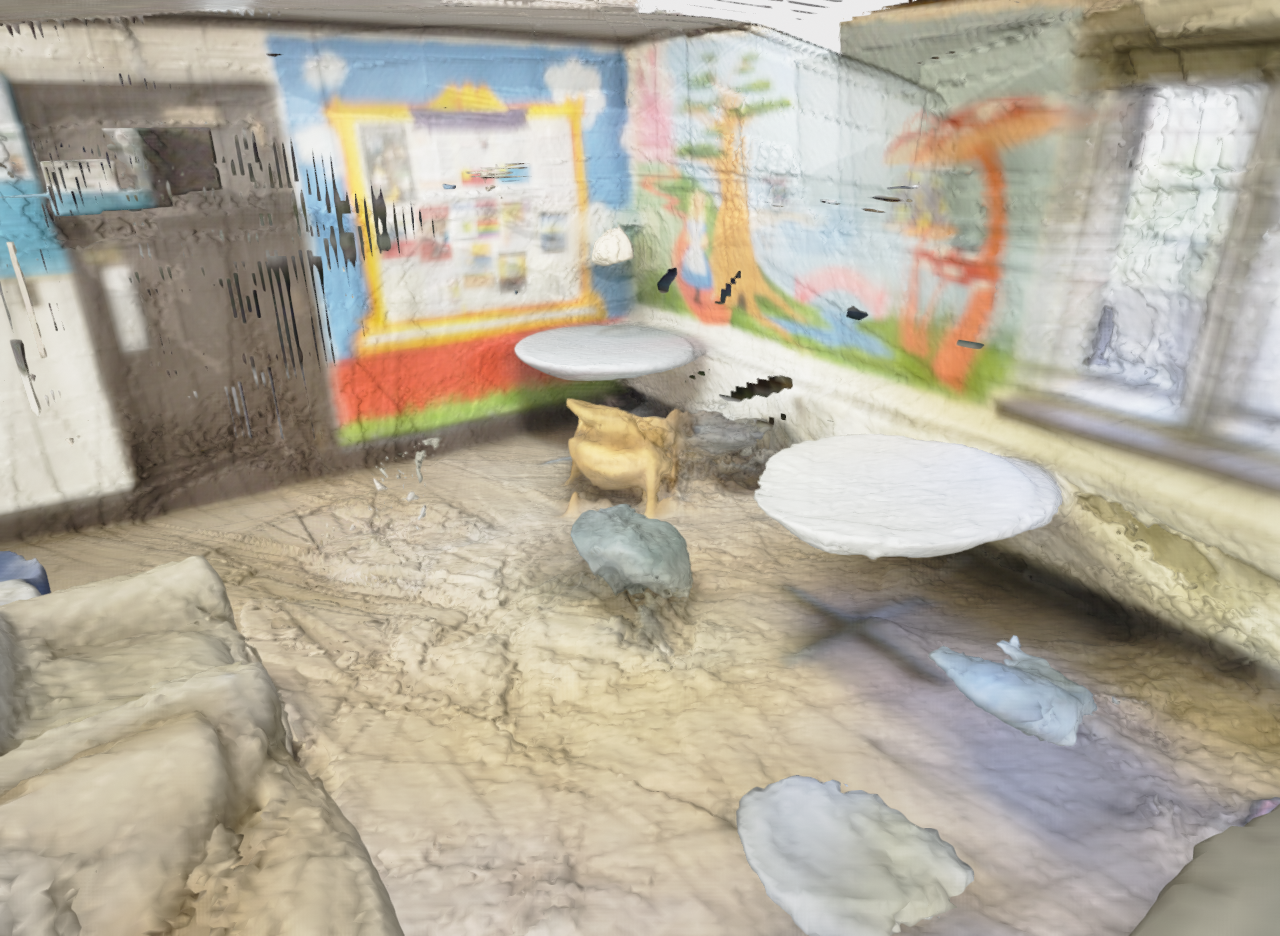} &
    \includegraphics[width=.48\linewidth]{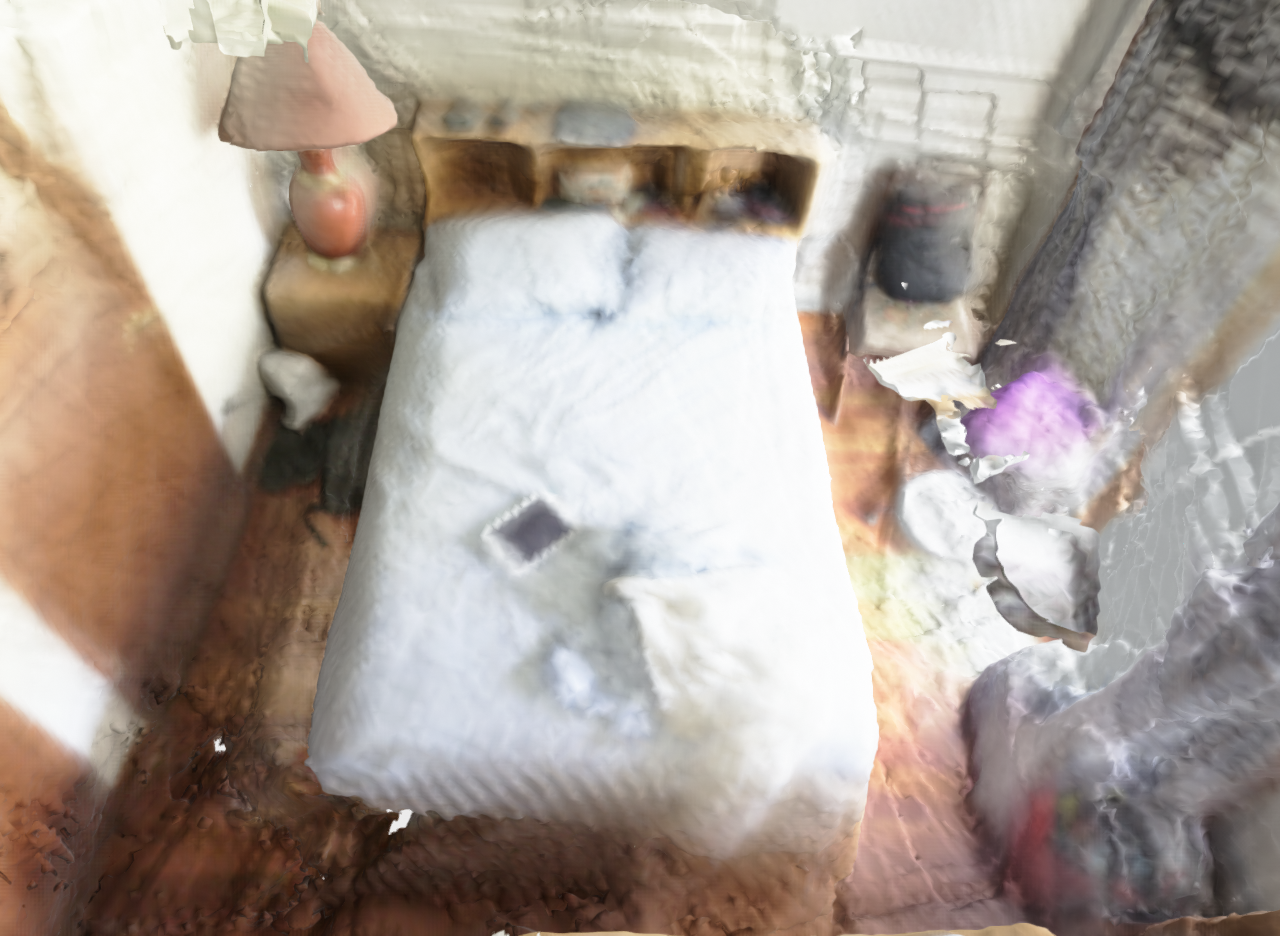}
    \end{tabular}
    \caption{With scale calibration and volumetric fusion, room-scale geometry initialization can be achieved from monocular depth without any optimization {of the voxel grid parameters}. The remaining task would be refining noisy regions and prune outliers.}
    \label{fig:init}
\end{figure}

\subsubsection{Depth, Color, and Semantic Fusion}
Following classical volumetric fusion scheme~\cite{newcombe2011kinectfusion, niessner2013real}, we project voxels $\vv$ back to the images to setup voxel-pixel associations, and directly optimize voxel-wise properties, namely SDF ($\theta_d$), color ($\theta_c$), and semantic label logits ($\theta_s$).

{
\begin{align}
  \theta_d(\vv) &= \argmin_{d} \sum_i \bigg( - \big(d_{\vv_{\to i}} - \dD_i(\pp_{\vv_{\to i}})\phi_i(\pp_{\vv_{\to i}})\bigg)^2, \\
  \theta_c(\vv) &= \argmin_{\cc} \sum_i \lVert \cc - \iI_i(\pp_{\vv_{\to i}}) \rVert^2, \\
  \theta_s(\vv) &= \frac{\sss^*}{\lVert \sss^* \rVert}, \sss^*=\argmin_{\sss} \sum_i \lVert \sss - \sS_i(\pp_{\vv_{\to i}}) \rVert^2,\hspace{-1mm}~\label{eq:semantic_mean}
\end{align}
}%
where the projection $\vv \to i$ is given by Eq.~\ref{eq:reproj_3d}. Note by definition, only associations with SDF smaller than a truncation bound will be considered, minimizing the effect of occlusion.
It is worth mentioning that we use a simple L2 loss for semantic logit instead of entropy losses, as it is considered one of the best practices in label fusion~\cite{mccormac2018fusion++}. The closed-form solutions of aforementioned voxel-pixel association losses are simply averages. Therefore, with minimal processing time, we can already achieve reasonable initial surface reconstruction by classical volumetric SDF and color fusion, see Fig.~\ref{fig:init}.

%% file: tex/method_refinement.tex
\begin{figure}[ht]
    \centering
    \footnotesize
    \begin{tabular}{@{}c@{}c@{}c@{}}
    \includegraphics[width=.33\columnwidth]{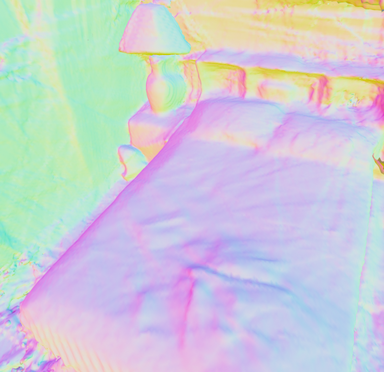} &
    \includegraphics[width=.33\columnwidth]{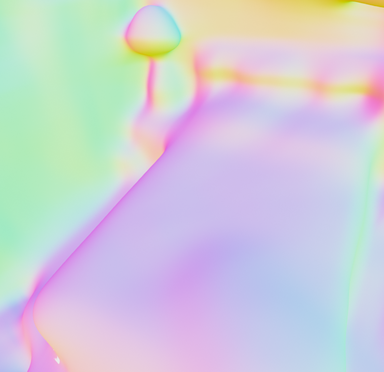} &
    \includegraphics[width=.33\columnwidth]{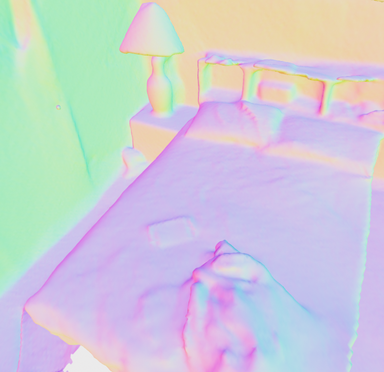} \\
    \includegraphics[width=.33\columnwidth]{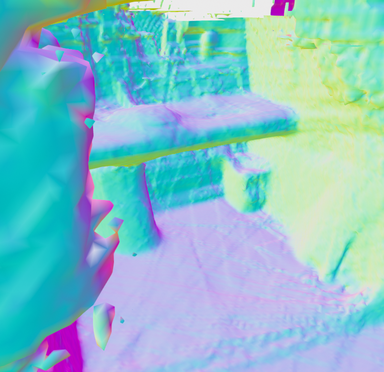} &
    \includegraphics[width=.33\columnwidth]{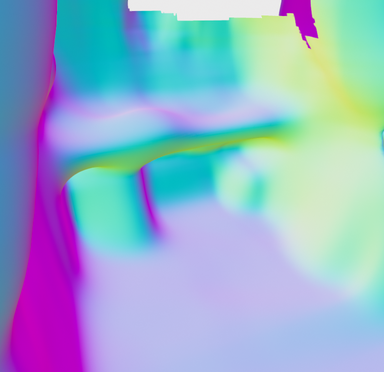} &
    \includegraphics[width=.33\columnwidth]{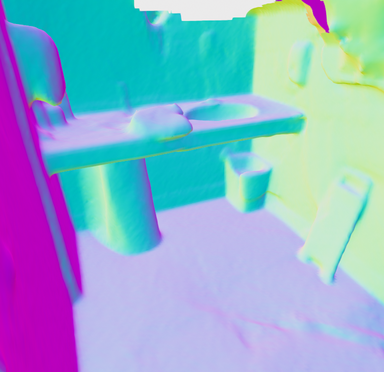} \\    
    (a) Init fusion & (b) De-noised & (c) Refined
    \end{tabular}
    \caption{Comparison between 3 stages of reconstruction: initialization, de-noising, and volume rendering refinement.}
    \label{fig:progress}
\end{figure}

\subsubsection{De-noising}
Direct fused properties, although being smoothed average of observations across frames \emph{per voxel}, are spatially noisy and can result in ill-posed SDF distributions along rays. Therefore, we perform a Gaussian blurring for the voxels along all the properties.
Thanks to the direct representation, with a customized forward sparse convolution followed by a property assignment, we could accomplish the filtering without backward optimizations. The effect of the de-noising operation can be observed in Fig.~\ref{fig:progress}(a)-(b).

\subsection{Differentiable Geometry Refinement}\label{subsec:refine}
\subsubsection{Volume Rendering}\label{subsubsect:rendering}
We follow MonoSDF~\cite{yu2022monosdf} to refine geometry using monocular priors. For a pixel $\pp$ from frame $i$, we march a ray $\xx(t) = \rr_o + t\cdot \rr_d$ to the sparse voxel grid, sample a sequence of points $\{\xx_k = \xx(t_k)\}$, apply volume rendering, and minimize the color, depth, and normal losses respectively:
{
\begin{align}
  \lL_c({\theta_c}, {\theta_d}) &= \bigg\lVert \sum_{k} w(\xx_k) f_{\theta_c}(\xx_k) - \iI_i(\pp) \bigg\rVert,\\
  \lL_d(\theta_d) &= \bigg \lVert \sum_{k} w(\xx_k) t_k - (a \dD_i(\pp) + b) \bigg\rVert^2 , \\
     \lL_n(\theta_d) &= \bigg \lVert \sum_k w(\xx_k) \nabla f_{\theta_d}(\xx_k) - \nN_i(\pp) \bigg\rVert, \\
  w(\xx_k) &= \exp \big(-\sum_{j < k} \alpha(\xx_j) \delta_j\big) \big(1 - \exp (-\alpha(\xx_k) \delta_k)\big),
\end{align}
}%
where $\delta_i = t_{i+1} - t_i$, depth scale $a$ and shift $b$ are estimated per minibatch in depth loss with least squares~\cite{ranftl2021vision}, and the density $\alpha(\xx_k) = l(f_{\theta_d}(\xx_k))$ is converted from SDF with a Laplacian density transform from VolSDF~\cite{yariv2021volume}. To accelerate, points are sampled in the sparse grid where valid voxel blocks have been allocated, and the empty space is directly skipped.
\subsubsection{Regularization}
Eikonal regularization~\cite{atzmon2020sal} forces SDF gradients to be close to 1, 
\begin{align}
  \lL_{\mathrm{Eik}} = \left(\lVert \nabla f_{\theta_d}(\xx)  \rVert - 1\right)^2.
\end{align}
Similar to related works~\cite{yariv2021volume,yu2022monosdf}, $\{\xx\}$s are samples combined with ray-based samples around surfaces, and uniform samples in the sparse grids. It is worth noting that in an explicit SDF voxel grid, $f_{\theta_d}$ and $\nabla f_{\theta_d}$ can be jointly computed in the same pass:
\begin{align}
  f_{\theta_d}(\xx) &= \sum_{\xx_i \in \mathrm{Nb}(\xx)} r(\xx, \xx_i) \theta_d(\xx_i), \\
  \nabla_{\xx} f_{\theta_d}(\xx) &= \sum_{\xx_i \in \mathrm{Nb}(\xx)} \nabla_\xx r(\xx, \xx_i) \theta_d(\xx_i),\label{eq:eikonal}
\end{align}
where $\theta_d(\xx_i)$ are directly stored SDF values at voxel grid points $\xx_i$, and $r$ is the trilinear interpolation ratio function that is a polynomial with closed-form derivatives. This circumvents costly double backward pass for autodiff gradient estimation~\cite{yariv2021volume,yu2022monosdf}, therefore speeds up training both by reducing computation burden and allowing larger batch size.

\subsubsection{Differentiable Continuous Semantic CRF}\label{subsubsec:crf}
Through differentiable volume rendering, we have achieved fine geometry reconstructions. 
We want to further sharpen the details at the boundaries of objects (\eg, at the intersection of a cabinet and the floor).
We resort to CRFs for finetuning all the properties, including colors, normals, and labels.
Unlike conventional CRFs where energy functions are defined on discrete nodes, we propose to leverage our data structure and devise a continuous CRF to integrate energy over the surface
\begin{equation}
    E(\mathbb{S}) = \int_\mathbb{S} \psi_u(\mathbf{x}) d\mathbf{x} + \int_\mathbb{S}\int_\mathbb{S} \psi_p(\mathbf{x}_i, \mathbf{x}_j) d\mathbf{x}_i d\mathbf{x}_j,
\end{equation}
where %$\mathbb{S}$ denotes the surface and 
$\mathbf{x} \in \mathbb{S}$ denotes a point on the surface. $\psi_u$ and $\psi_p$ denote unary and pairwise Gibbs energy potentials.
Following Krahenbuhl~\etal~\cite{krahenbuhl2011efficient}, we adopt the Gaussian edge potential
\begin{equation}
    \psi_p(\mathbf{x}_i, \mathbf{x}_j) = \mu_\mathrm{prop}(\mathbf{x}_i, \mathbf{x}_j) \exp\left(-(\mathbf{f}_i - \mathbf{f}_j)^T\Lambda(\mathbf{f}_i - \mathbf{f}_j)\right),
\end{equation}
where $\mu_\mathrm{prop}$ denotes a learnable compatibility function of a node property (\eg~normal), $\exp$ computes the consistency strength between nodes from feature distances with
the precision matrix $\Lambda$.
A feature $\ff_i$ concatenates 3D positions, colors, normals, and label logits queried at $\xx_i$. We approximate the integration over the surface with Monte Carlo sampling by finding zero-crossings from random camera viewpoints. 

The variational inference of the Gibbs energy potential with the mean-field approximation results in a simple update equation
\begin{equation}\label{eq:vi_update}
  \scalemath{0.95}{Q(\mathbf{x}_i)^+ \propto \exp\left( - \psi_u(\mathbf{x}_i) - \sum_j \psi_p(\mathbf{f}_i, \mathbf{f}_j)Q(\mathbf{x}_j) \right).}
\end{equation}
Note that the summation in Eq.~\ref{eq:vi_update} is over all the sample points and computationally prohibitive. Thus, we use a high-dimensional permutohedral lattice convolution~\cite{adams2010fast} to accelerate the message passing, driven by our collision-free hash map at high dimensions. 

For each of the target properties $\mathrm{prop} \in \{\mathrm{color}, \mathrm{normal}, \mathrm{label}\}$, we define a loss $\lL_\mathrm{prop} = D_f\Big(\xx_\mathrm{prop} \lVert Q(\xx_\mathrm{prop}) \Big)$ with f-diveregence, conditioned on the remaining properties plus the 3D positions. A joint loss is defined to optimize all the properties:
\begin{align}
    &\lL_\mathrm{CRF} = \lambda_{\mathrm{color}}\lL_\mathrm{CRF}^\mathrm{color} + \lambda_{\mathrm{normal}}\lL_\mathrm{CRF}^\mathrm{normal} + \lambda_{\mathrm{label}}\lL_\mathrm{CRF}^\mathrm{label}. \label{eq:crf}
\end{align}

\subsubsection{Optimization}
The overalls loss function at refinement stage is
\begin{align}
    \lL = \lL_c + \lambda_d \lL_d + \lambda_n \lL_n + \lambda_\textrm{Eik} \lL_\textrm{Eik} + \lL_\textrm{CRF}.
\end{align}
We optimize the grid parameters $\{\theta_d, \theta_c, \theta_s\}$ with RMSProp starting with a learning rate $10^{-3}$, and an exponential learning rate scheduler with $\gamma=0.1$.

%% file: tex/experiment.tex
\subsection{Setup}
We follow Manhattan SDF~\cite{guo2022neural} and evaluate on 4 scenes from ScanNet~\cite{dai2017scannet} and 4 scenes from 7-scenes~\cite{glocker2013real} in evaluation. We use reconstruction's F-score as the major metric, along with distance metrics (accuracy, completeness), precision, and recall.
We compare against COLMAP~\cite{schonberger2016structure}, NeRF~\cite{mildenhall2021nerf}, UNISURF~\cite{oechsle2021unisurf}, NeuS~\cite{wang2021neus}, VolSDF~\cite{yariv2021volume}, Manhattan SDF~\cite{guo2022neural}, and MonoSDF~\cite{yu2022monosdf}. We train MonoSDF to obtain output mesh. For the rest of the compared approaches, we reuse reconstructions provided by the authors from Manhattan SDF~\cite{guo2022neural}, and evaluate them against high-resolution ground truth via TSDF fusion. The evaluation metric and implementation details are in supplementary.

For geometric priors, unlike MonoSDF~\cite{yu2022monosdf} that generates monocular cues from $384 \times 384$ center crops, we follow DPT~\cite{ranftl2021vision}'s resizing protocol and adapt Omnidata~\cite{eftekhar2021omnidata} to obtain $480\times 640$ full resolution cues. 

In all the experiments, we use a $8^3$ voxel block grid with a voxel size $1.5$cm. At each step, we randomly sample $1024$ rays per image with a batch size of $64$.
Due to the reasonable geometric initialization, the loss usually drops drastically within $2 \times 10^3$ iterations, and converges at $10^4$ iterations, therefore we terminate training at $10^4$ steps for all scenes. Thanks to the efficient data structure, accelerated ray marching, and closed-form SDF gradient computation, it takes less than 30 mins to reconstruct a scene on a mid-end computer with an NVIDIA RTX 3060 GPU and an Intel i7-11700 CPU.
\input{table/speed.tex}
\begin{figure}[t]
    \centering
    \includegraphics[width=\columnwidth]{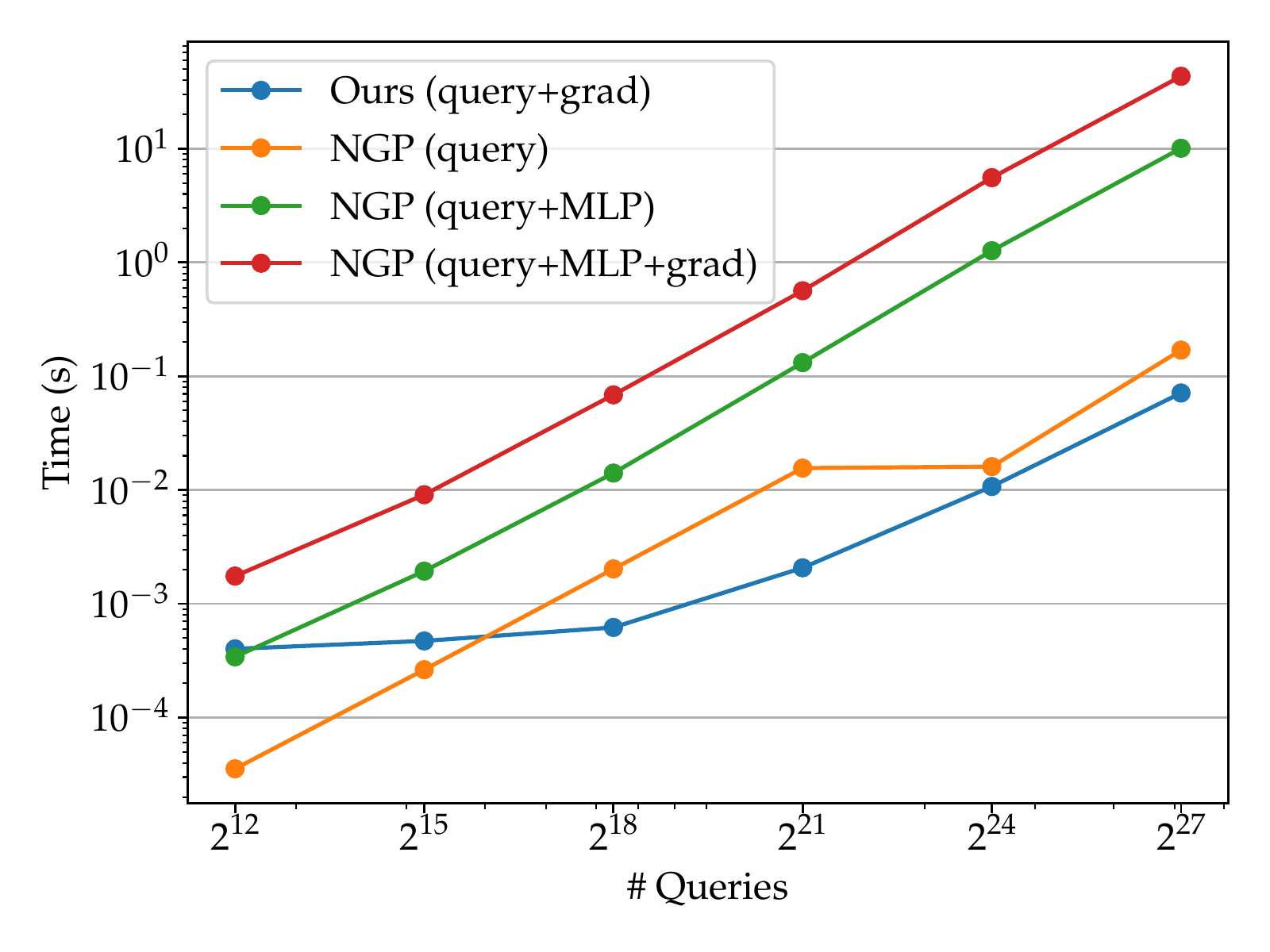}
    \caption{Query time comparison between ours and NGP-grid, lower is better. For end-to-end query, ours is two magnitudes faster, and maintains a high efficiency with a large number of point query. For the grid query operation itself, ours also have a better performance than multiresolution feature grids.}
    \label{fig:query_time}
\end{figure}

\subsection{Runtime Comparison}
We first profile the SDF query time given a collection of points on the aforementioned machine. Specifically, we sample $k^3, k \in \{2^4, \cdots, 2^9\}$ grid points of 3D resolution $k$, query the SDF and their gradients, and compare the run time. This is frequently used for Marching Cubes~\cite{lorensen1987marching} (requiring SDFs) and global Eikonal regularization~\cite{atzmon2020sal} (requiring SDF gradients).
We compare against MonoSDF's NGP-grid backbone that uses the multi-resolution grid from Instant-NGP~\cite{muller2022instant}. In this implementation, three steps are conducted to obtain required values: query from the feature grid; SDF inference from an MLP; SDF grad computation via autograd. In contrast, ours allows its explicit computation in one forward pass, see Eq.~\ref{eq:eikonal}. Fig.~\ref{fig:query_time} shows the breakdown time comparison.
\input{table/accuracy.tex}

We also show the training and inference time comparison in Table~\ref{table:recon-accuracy}. Due to the fine initialization and sparse data structure with accelerated ray sampling, our approach can complete training in less than half an hour, and allows fast rendering of color and depth at inference time.

\subsection{Reconstruction Comparison}
The comparison of reconstruction accuracy can be seen in Table~\ref{table:recon-accuracy}. We can see that our approach achieves high accuracy at initialization that surpasses various baselines. With volume rendering and CRF refinement, it reaches comparable accuracy to the state-of-the-art MonoSDF~\cite{yu2022monosdf} on ScanNet scenes, and achieves better results on 7-scenes. The last three rows serve as the ablation study, showing a major gain from volume rendering followed by a minor refinement gain from CRF.

\begin{figure}[ht]
    \centering
    \footnotesize
    \begin{tabular}{@{}c@{}c@{}c@{}}
    \includegraphics[width=.33\columnwidth]{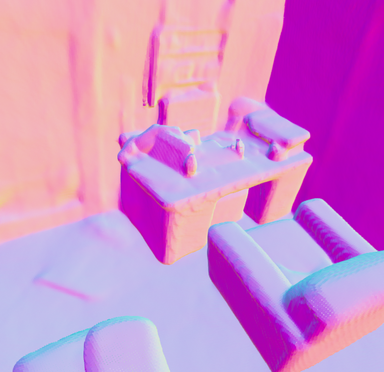} &
    \includegraphics[width=.33\columnwidth]{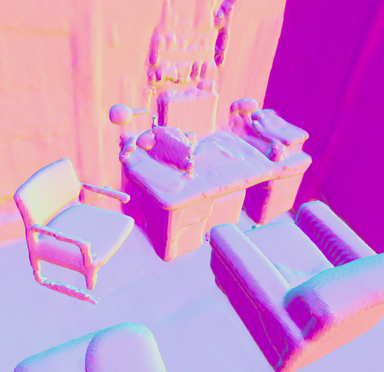} &
    \includegraphics[width=.33\columnwidth]{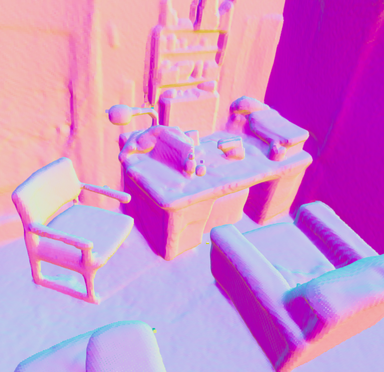} \\
    \includegraphics[width=.33\columnwidth]{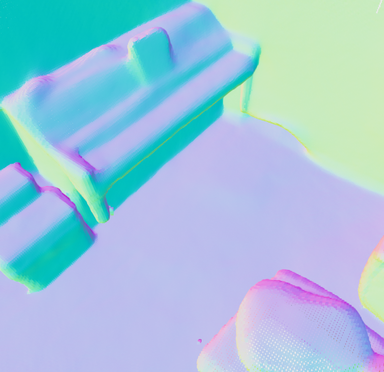} &
    \includegraphics[width=.33\columnwidth]{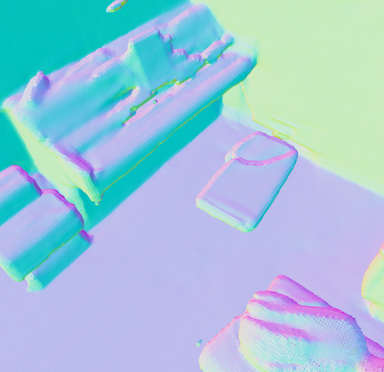} &
    \includegraphics[width=.33\columnwidth]{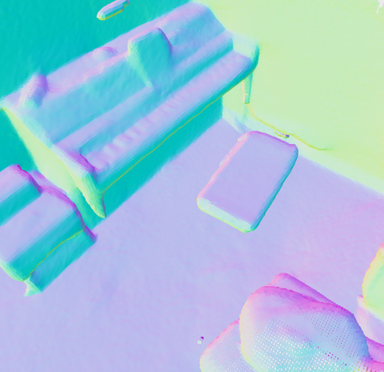} \\   
    {(a) MonoSDF-MLP} & {(b) MonoSDF-Grid} & {(c) Ours}
    \end{tabular}
    \caption{Detail comparisons between our method and current state-of-the-art neural implicit method MonoSDF~\cite{yu2022monosdf}. We preserve better geometry details while being faster.}
    \label{fig:details-0050}
\end{figure}

We also demonstrate qualitative scene-wise geometric reconstruction in Fig.~\ref{fig:qualitative-0050}, and zoomed-in details in Fig.~\ref{fig:details-0050}. It is observable that while achieving similar global completeness, our method enhances details thanks to the adaptive voxel grid and direct SDF mapping from coordinates to voxels.
\vspace{-1mm}

\begin{figure}[t]
    \centering
    \includegraphics[width=\columnwidth]{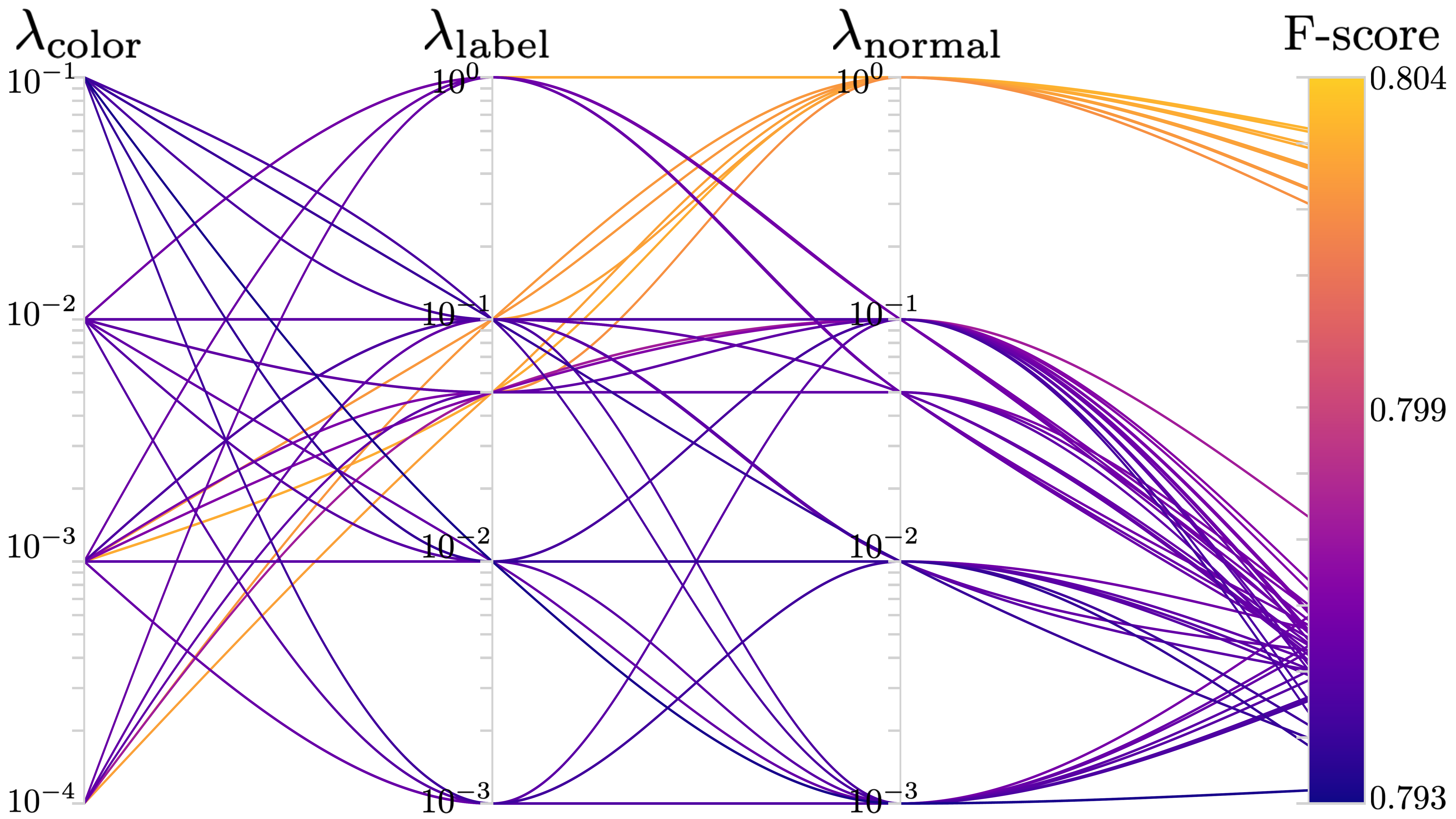}
    \caption{Control experiments of the CRF modules' impact to final reconstruction quality on scene 0084, see~Eq.~\ref{eq:crf}.}
    \label{fig:crf-ablation}
\end{figure}

The control experiments of CRF's incorporated properties are visualized in Fig.~\ref{fig:crf-ablation}, where we see that semantic labels and normals have the highest impact on reconstruction quality. Colors, on the other hand, have a lower impact mostly due to the prevalent appearance of motion blurs and exposure changes in the benchmark dataset.
%\input{table/crf.tex}
% \subsection{Limitations and Future Work}
The same reason also affects feature-based SfM and monocular depth estimate and leads to reduced performance of our approach on certain sequences, see supplementary.
We plan to incorporate more advanced semi-dense reconstruction~\cite{teed2021droid,teed2022deep} for robust depth prior estimate.

%% file: table/speed.tex
\begin{table}[htbp]
  \centering
\caption{Train and inference time (per image) analysis on the ScanNet scene 0084. Our approach both trains and evaluates faster.}
\resizebox{0.85\columnwidth}{!}{
  \begin{tabular}{l cc}
  \toprule
  Method         & Train (h) & Inference (s) \\
  \midrule
  NeuS~\cite{wang2021neus}            & 6.64       & 28.32         \\
  VolSDF~\cite{yariv2021volume}       & 8.33      & 29.64         \\
  ManhattanSDF~\cite{guo2022neural}   & 16.68      & 28.49        \\
  MonoSDF (MLP)~\cite{yu2022monosdf}  & 9.89       & 33.80         \\
  MonoSDF (Grid)~\cite{yu2022monosdf} & \two{4.36}       & \two{19.13}         \\
  \midrule
  Ours           & \one{0.47}       & \one{0.25}          \\
\bottomrule
\end{tabular}
}
\end{table}

%% file: table/accuracy.tex
\begin{table*}[htbp]
  \centering
  \caption{Quantitative comparison of reconstruction quality. While being much faster, our approach is comparable to the state-of-the-art MonoSDF~\cite{yu2022monosdf} on ScanNet~\cite{dai2017scannet} and better on 7-scenes~\cite{glocker2013real}.
  %\orl{missing second best for Comp column}
  }
  \resizebox{.99\textwidth}{!}{
\begin{tabular}{l ccccc | ccccc}
  \toprule
  \multirow{2}{*}{Method} & \multicolumn{5}{c}{ScanNet} & \multicolumn{5}{c}{7-Scenes} \\
  \cmidrule(lr){2-11}
  & Acc $\downarrow$ & Comp $\downarrow$ & Prec $\uparrow$ & Recall $\uparrow$  & F-score $\uparrow$  & Acc $\downarrow$ & Comp $\downarrow$ & Prec $\uparrow$  & Recall $\uparrow$  & F-score $\uparrow$  \\
  \midrule
  COLMAP~\cite{schonberger2016structure}        & 0.074 & 0.239 & 0.602 & 0.363 & 0.442 & \one{0.069} & 0.417 & \one{0.536} & 0.202 & 0.289 \\
  NeRF~\cite{mildenhall2021nerf}          & 0.605 & 0.178 & 0.186 & 0.302 & 0.225 & 0.573 & 0.321 & 0.159 & 0.085 & 0.083 \\
  UNISURF~\cite{oechsle2021unisurf}       & 0.497 & 0.167 & 0.224 & 0.327 & 0.265 & 0.407 & 0.136 & 0.195 & 0.301 & 0.231 \\
  NeuS~\cite{wang2021neus}          & 0.166 & 0.221 & 0.296 & 0.237 & 0.262 & 0.151 & 0.247 & 0.313 & 0.229 & 0.262 \\
  VolSDF~\cite{yariv2021volume}        & 0.378 & 0.139 & 0.284 & 0.330 & 0.301 & 0.285 & 0.140 & 0.220 & 0.285 & 0.246 \\
  ManhattanSDF~\cite{guo2022neural}  & 0.081 & 0.099 & 0.626 & 0.544 & 0.581 & 0.112 & {0.133} & 0.351 & {0.326} & 0.336 \\
  MonoSDF (MLP)~\cite{yu2022monosdf} & \one{0.031} & {0.057} & \two{0.783} & {0.652} & {0.710} & \two{0.097} & 0.192 & \two{0.441} & 0.311 & {0.361} \\  
  MonoSDF (Grid)~\cite{yu2022monosdf} & \two{0.034} & \one{0.046} & \one{0.796} & \one{0.711} & \one{0.750} & 0.113 & {0.100} & {0.433} & {0.392} & {0.411} \\ 
  \midrule
  Ours (Scale Optim.)  & 0.058 & 0.064 &  0.655 & 0.605 & 0.627 & 0.151 & \two{0.080}& 0.367 & \two{0.462} & 0.409 \\
  Ours (+ Volume Rendering) & {0.045} & {0.060} & {0.774} & {0.667}     & \two{0.714} & 0.140 & {0.081} & 0.417 & {0.450} & \two{0.433}\\
  Ours (+ CRF) & {0.042} & \two{0.056} & {0.751} & \two{0.678} & {0.710}     & {0.136} & \one{0.079} & {0.436} & \one{0.475}& \one{0.454} \\
\bottomrule
\end{tabular}
}
\label{table:recon-accuracy}
\end{table*}

%% file: tex/conclusion.tex
We propose an efficient monocular scene reconstruction system. Without an MLP, our model is built upon a differentiable globally sparse and locally dense data structure allocated around approximate surfaces.
We develop a scale calibration algorithm to align monocular depth priors for fast geometric initialization, and apply direct refinement of voxel-level SDF and colors using differentiable rendering. 
We further regularize the voxel-wise properties with a high-dimensional continuous CRF that jointly refines color, geometry, and semantics in 3D.
Our method is $10\times$ faster in training and $100 \times$ faster in inference, while achieving similar reconstruction accuracy to state-of-the-art.

%% file: tex/supplement_sec.tex
\newcommand{\vcell}[1]{\begin{tabular}{l} #1 \end{tabular}}

\section*{Supplementary}
\subsection*{Depth Scaler Optimization}
Our system adopts monocular depth map predictions from off-the-shelf networks~\cite{eftekhar2021omnidata} using the DPT backbone~\cite{ranftl2021vision}. However, these depth priors are not metric and the scale of each depth prediction is independent of others. Thus, we define the unary and binary (pairwise) constraints to estimate consistent metric scales.

\subsubsection*{Unary Constraints}
Our pipeline relies on COLMAP's~\cite{schonberger2016structure} sparse reconstruction for unary constraints.
COLMAP supports sparse reconstruction with or without poses. 
Both modes start with SIFT~\cite{lowe2004distinctive} feature extraction and matching. 
The \emph{with pose} mode then runs triangulation, while the \emph{without pose} mode runs bundle adjustment to also estimate poses.
\emph{With pose} mode usually runs within 1 min, while the \emph{without pose} mode often finishes around 5 mins for a sequence with several hundred frames. 
While our system integrates both modes, for fair comparison on the benchmark datasets, we adopt the \emph{with pose} mode in quantitative experiments where ground truth poses from RGB-D SLAM are given.
Fig.~\ref{fig:sparse_recon} shows the sparse reconstructions from the \emph{with pose} mode.

% We use the sparse reconstruction to construct unary estimates for monocular depth scales.
\subsubsection*{Binary Constraints}
\begin{figure*}[t]
    \centering
    \footnotesize
    \begin{tabular}{c@{}c@{}c@{}c}
    \includegraphics[width=.24\textwidth]{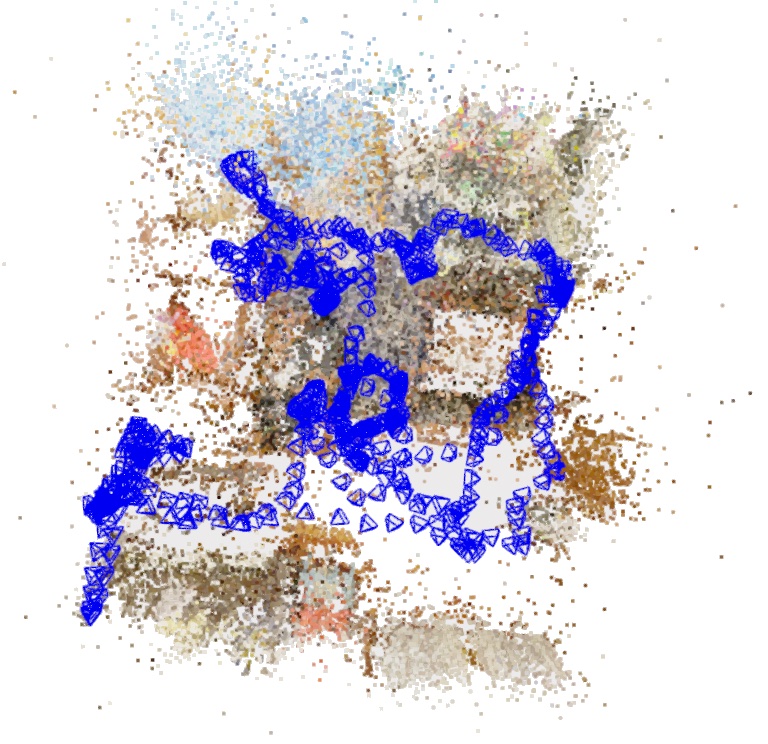} &  
    \includegraphics[width=.24\textwidth]{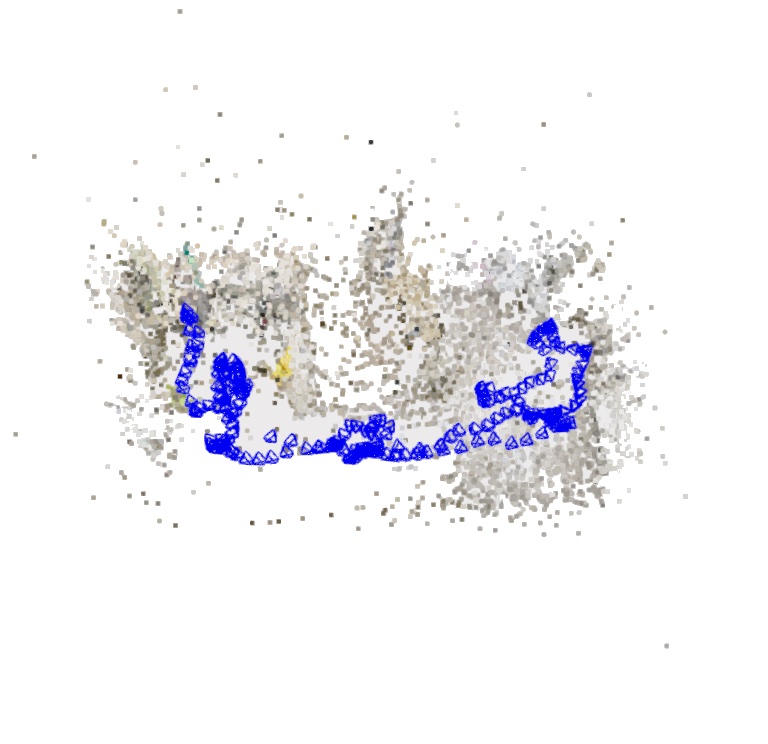} &
    \includegraphics[width=.24\textwidth]{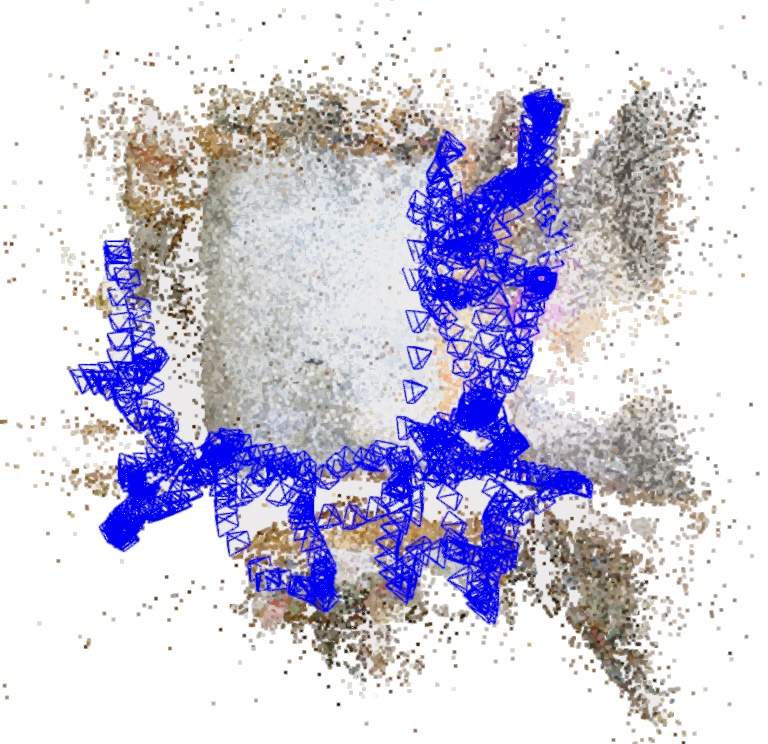} &  
    \includegraphics[width=.24\textwidth]{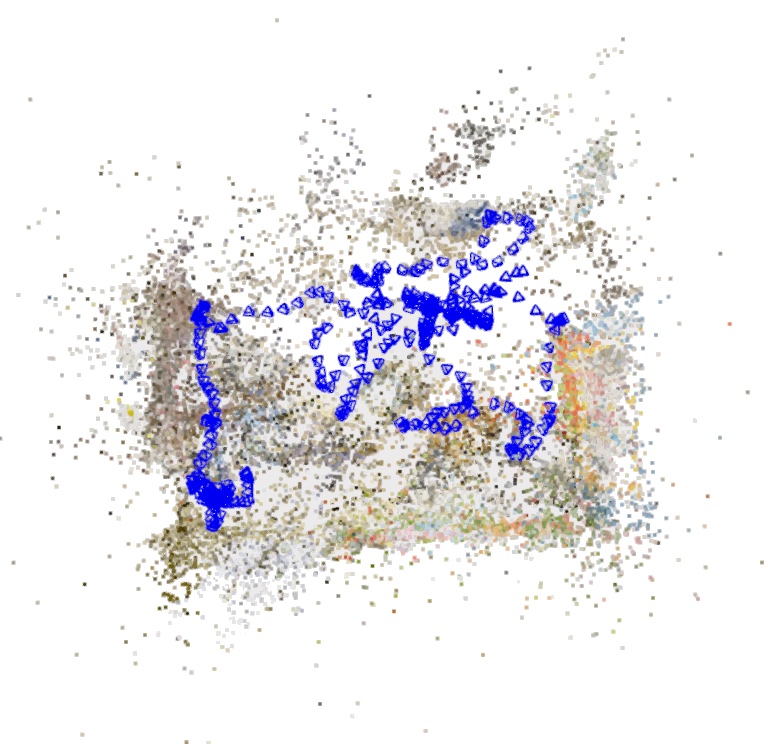} \\
    \includegraphics[width=.24\textwidth]{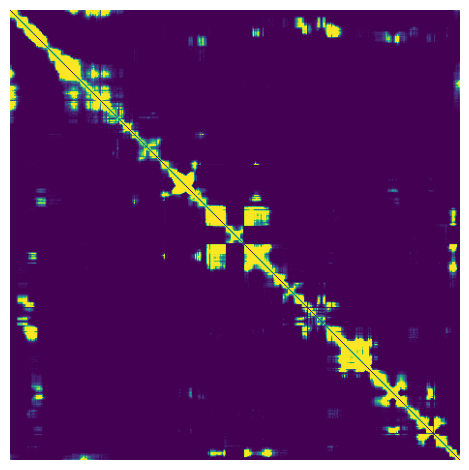} &  
    \includegraphics[width=.24\textwidth]{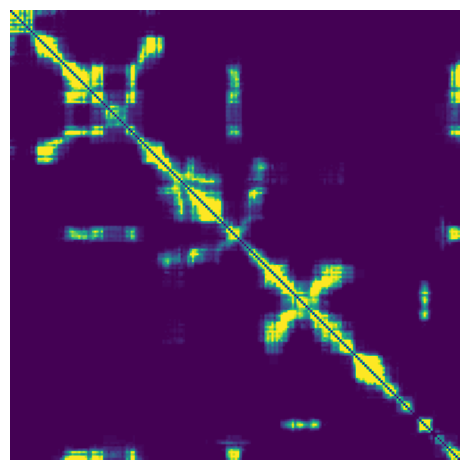} &
    \includegraphics[width=.24\textwidth]{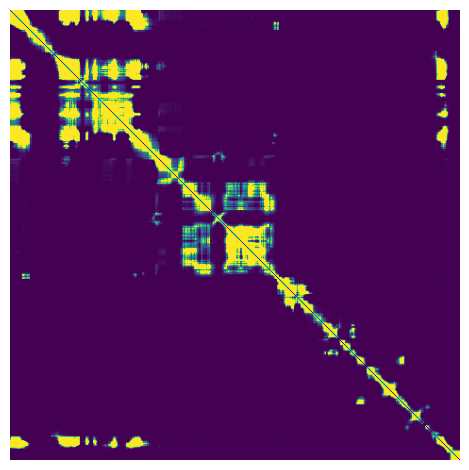} &  
    \includegraphics[width=.24\textwidth]{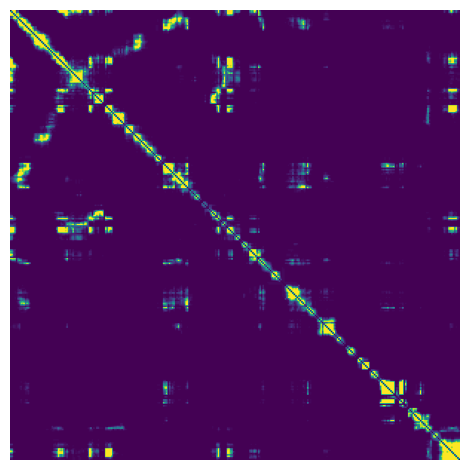} \\    
    0050 & 0084 & 0580 & 0616
    \end{tabular}
    \caption{Sparse reconstruction and covisibility matrix of ScanNet scenes selected by ManhattanSDF~\cite{guo2022neural}.}
    \label{fig:sparse_recon}
\end{figure*}

Once we have camera poses and the sparse reconstruction, we can define which triangulated feature points are visible to which cameras (covisible). Thus, we can create pairwise reprojection constraints between frames, similar to loop closures in the monocular SLAM context~\cite{mur2015orb}.
% Covisibility is known to establish correspondences between frames that share features (\ie~co-visible features between frames), especially in the SLAM context~\cite{mur2015orb}.
%
We directly retrieve the feature matches obtained by COLMAP, and setup such frame-to-frame covsibility constraints. Fig.~\ref{fig:sparse_recon} shows the covisibility matrices, where entry $(i, j)$ indicates the number of covisible features between frame $i$ and $j$. They are used to establish binary constraints between frames for refining monocular depth scales.

\subsection*{Volumetric Fusion}
Eq. 9 in the main paper shows the least squares to initialize voxel-wise SDF. The more detailed implementation follows KinectFusion~\cite{newcombe2011kinectfusion}, where a truncation function $\psi$ is used to reject associations.
\begin{align}
  \theta_d(\vv) &= \argmin_{d} \sum_i \lVert d - \psi \big(d^o, \mu\big) \rVert^2,\\
  d^o &= d_{\vv_{\to i}} - \dD_i(\pp_{\vv_{\to i}})\phi_i(\pp_{\vv_{\to i}}), \\
  \psi(x, \mu) &= \min(x, \mu),
\end{align}
where $\mu$ is the truncation distance. $\mu$ is associated with the \emph{Dilate} operation and voxel block resolution in Eq. 7-8 in the main paper. 
Formally, we define
\begin{align}
    \mathrm{Dilate}_R(\xx) = \bigg\{ \xx_i \mid \bigg\lVert \xx_i - \left\lfloor \frac{\xx}{L} \right\rfloor \bigg\rVert_0 \le R \bigg\},
\end{align}
where $L$ is the voxel block size, $\xx_i$ are quantized grid points around, and $R$ is the dilation radius. We use $R=2$ (corresponding to two $8^3$ voxel blocks) to account for the uncertainty around surfaces from the monocular depth prediction. Correspondingly, we use $\mu = L \cdot R$ to truncate the SDF.

The volumetric fusion runs at $50$ Hz with RGB and SDF fusion, and at $30$ Hz when additional semantic labels are also fused, hence serves as a fast initializer.

\subsection*{Hyper Parameters}
We followed~\cite{yu2022monosdf}'s hyperparameter choices and used $\lambda_{d} = 0.1, \lambda_{n}=0.05$ for the rendering loss. 

For regularizors, we obtained from hyper param sweeps from the 0084 scene of ScanNet that $\lambda_{\mathrm{eik}}=0.1$ for the Eikonal loss, and $\lambda_{\mathrm{color}}=10^{-3}, \lambda_{\mathrm{label}}=0.1, \lambda_{\mathrm{normal}}=1$ for the CRF loss. 

In Gaussian kernels, we fix $\sigma_{\mathrm{sdf}} = 1.0$ and $\sigma_{\mathrm{color}} = 0.1$.
%}
\subsection*{Evaluation}
\subsubsection*{Metrics}
We follow the evaluation protocols defined by ManhattanSDF~\cite{guo2022neural}, where the metrics between predicted point set $P$ and ground truth point set $P^*$ are
{\small
\begin{align}
    D(p, p^*) &= \lVert p -  p^*\rVert,\\
    \mathrm{D}_\mathrm{Acc}(P, P^*) &= \mean_{p \in P} \min_{p^* \in P^*} D(p, p^*), \\
    \mathrm{D}_\mathrm{Comp}(P, P^*) &= \mean_{p^* \in P^*} \min_{p \in P} D(p, p^*),\\
    \mathrm{Prec}(P, P^*) &= \mean_{p \in P}\Big( \big(\min_{p^* \in P^*}  D(p, p^*) \big) < T\Big), \\
    \mathrm{Recall}(P, P^*) &= \mean_{p^* \in P^*} \Big(\big(\min_{p \in P} D(p, p^*)\big) < T\Big) , \\
    \mathrm{F\textrm{-}score}(P, P^*) &= \frac{2 \cdot \mathrm{Prec} \cdot \mathrm{Recall}}{\mathrm{Prec} + \mathrm{Recall}},
\end{align}
}%
where $T=5$cm.

\subsubsection*{Generation of $P$ and $P^*$}
We follow previous works~\cite{guo2022neural, yu2022monosdf} that applied TSDF refusion to generate $P$ for evaluation: use Marching Cubes~\cite{lorensen1987marching} to generate a global mesh; render depth map from mesh at selected viewpoints to crop points out of viewports; apply TSDF fusion~\cite{zhou2018open3d} to obtain the final mesh and point cloud $P$. 
For fairness, we render depth at the resolution $480\times 640$ for all approaches to be consistent with input (in contrast to MonoSDF that uses $968 \times 1296$ in their released evaluation code), and conduct refusion to a voxel grid at the resolution of $1$cm. 

To ensure the same surface coverage, we generate ground truth $P^*$ at the same viewpoints with the same image and voxel resolution, only replacing rendered depth with ground truth depth obtained by an RGB-D sensor.

%For ScanNet~\cite{dai2017scannet}, we do not use the low resolution ground truth point cloud used by ManhattanSDF~\cite{guo2022neural}. Instead, we generate high resolution mesh via TSDF fusion at the voxel resolution of $1cm$ using Open3D's sparse voxel grid~\cite{zhou2018open3d, dong2021ash}. This choice 

\subsection*{Additional Experimental Results}
\subsubsection*{Ablation of scale optimization}

To further illustrate the necessity of per-frame scale optimization, we show quantitative reconstruction results without scale optimization in Table~\ref{table:scale-ablation}. Here, volumetric fusion is conducted on an estimated single scale factor across all frames between monocular depth and SfM, resulting in poor initial reconstruction.
\begin{table}[h]
\centering
\caption{Initial reconstruction results without per-frame scale optimization (\emph{c.~f.~} Ours (Init) in Table~\ref{table:recon-accuracy-scannet}-\ref{table:recon-accuracy-7scenes}.)}
\label{table:scale-ablation}
\resizebox{.49\textwidth}{!}{
\begin{tabular}{l|ccccc}
\hline
& Acc $\downarrow$ & Comp $\downarrow$ & Prec $\uparrow$ & Recall $\uparrow$ & F-score $\uparrow$\\
\hline
ScanNet & 0.42 & 0.19 & 0.13 & 0.28 & 0.17\\
7-Scenes & 0.36 & 0.12 & 0.19 & 0.43 & 0.26\\
\hline
\end{tabular}
}
\end{table}

\subsubsection*{Fusion and Refinement}
Please see \href{https://youtu.be/87guWiDZkSI}{video supplementary} for the incremental fusion from scaled depth, and the refinement stage that converges to general shapes within several hundred steps.

\subsubsection*{Scene-wise statistics on ScanNet}
\begin{figure*}[h]
    \centering
    \footnotesize
    \begin{tabular}{c@{}c@{}c@{}c}
    \includegraphics[width=.24\textwidth]{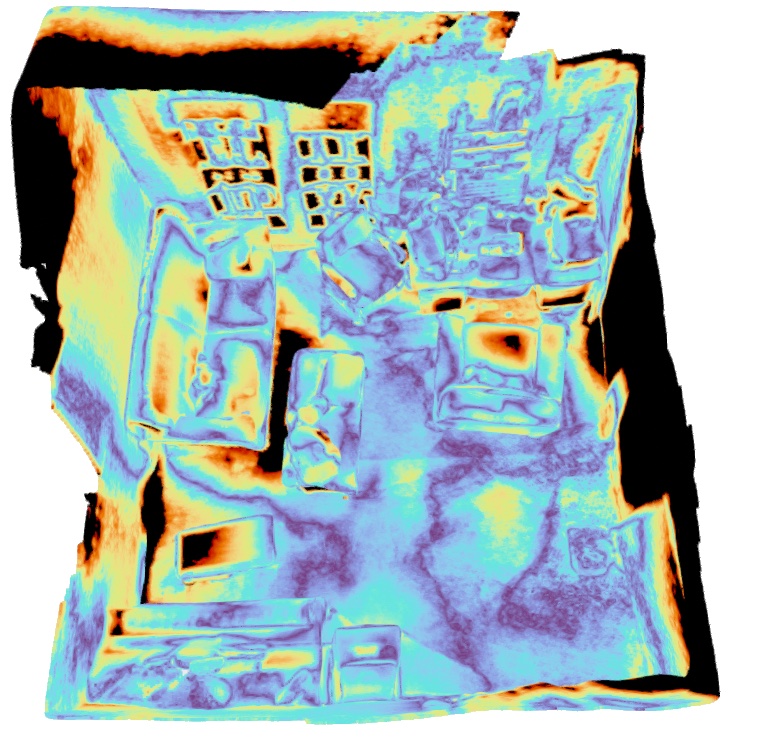} &  
    \includegraphics[width=.24\textwidth]{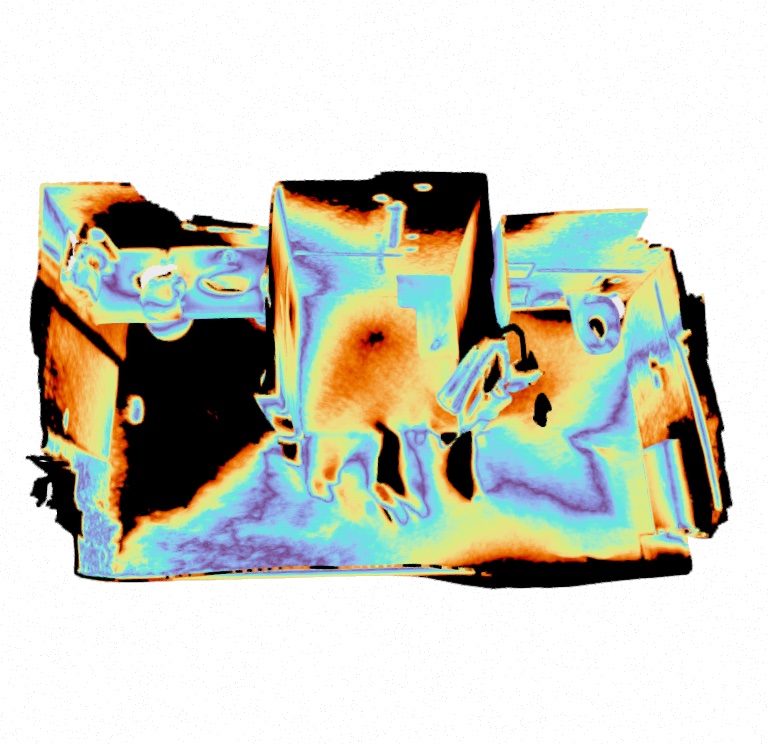} &
    \includegraphics[width=.24\textwidth]{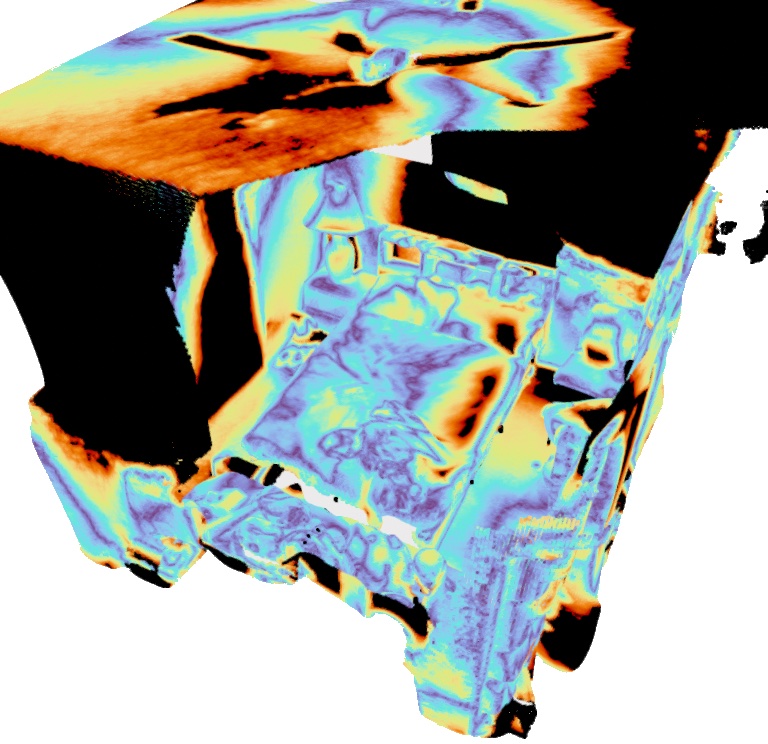} &  
    \includegraphics[width=.24\textwidth]{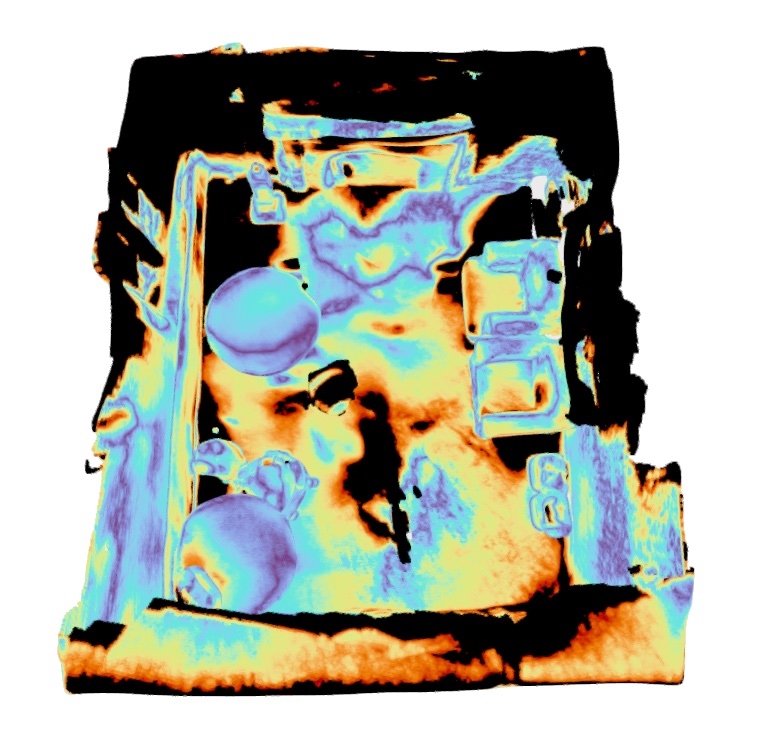} \\    
    \includegraphics[width=.24\textwidth]{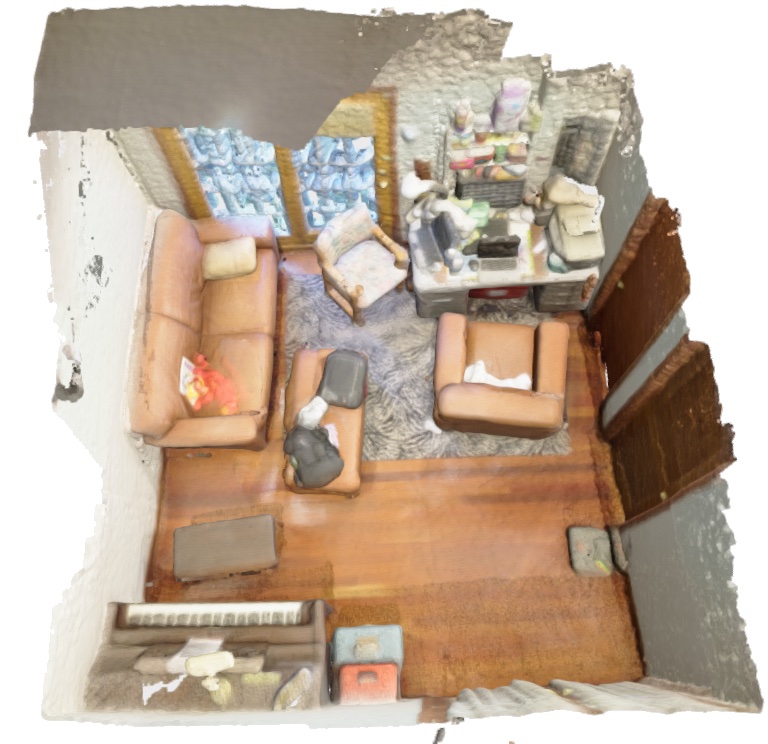} &  
    \includegraphics[width=.24\textwidth]{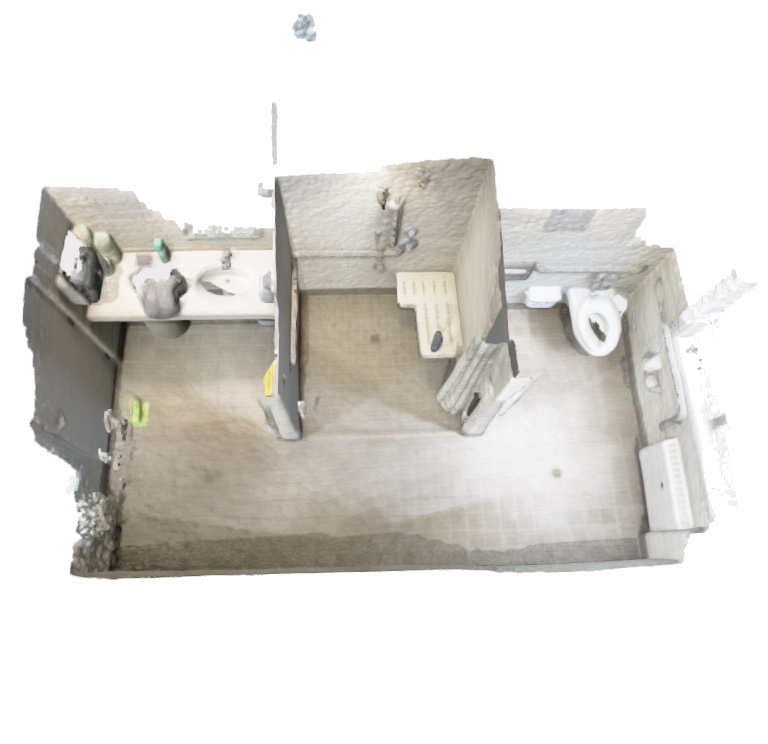} &
    \includegraphics[width=.24\textwidth]{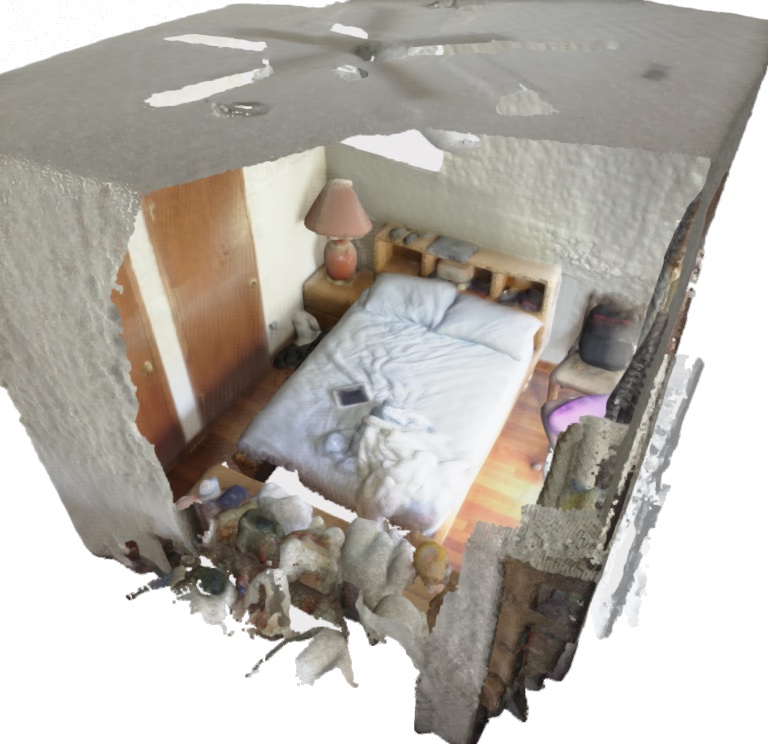} &  
    \includegraphics[width=.24\textwidth]{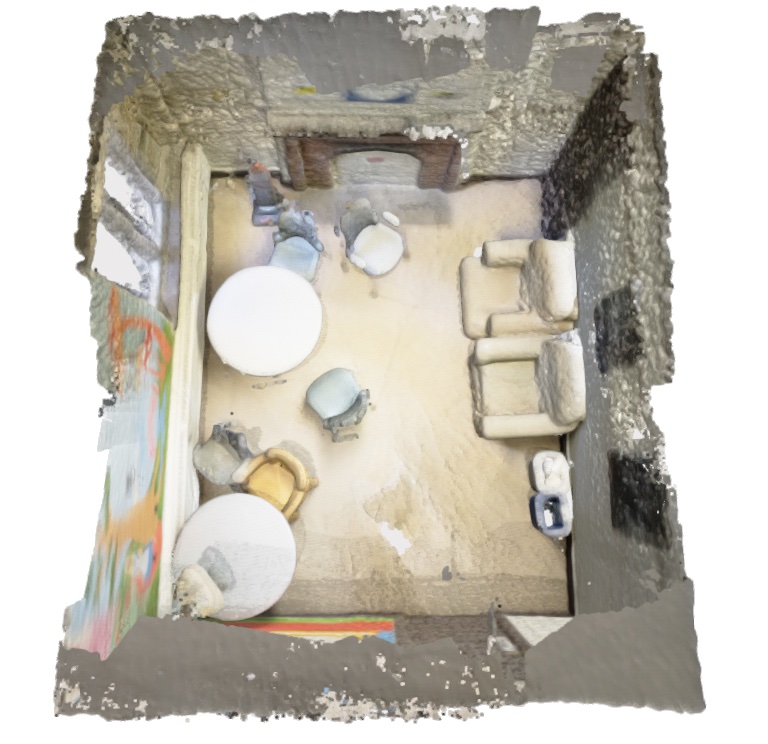} \\        
    0050 & 0084 & 0580 & 0616
    \end{tabular}
    \caption{Error heatmap from our reconstruction (first row) to groundtruth (second row) for each scene in ScanNet~\cite{dai2017scannet}. Points are colorized by distance error ranging from 0 (blue) to 5cm (red) to its nearest neighbor in ground truth. Points with error larger than 5cm are regarded as outliers and colored in black.}
    \label{fig:error_map}
\end{figure*}

We use reconstructed mesh provided by ManhattanSDF~\cite{guo2022neural}, and report scene-wise statistics in Table~\ref{table:recon-accuracy-scannet}. Reconstructions and corresponding ground truths are shown in Fig.~\ref{fig:error_map}.

It is observable that our reconstructions have low error at fine details with rich textures (\eg~0050, furniture in 0580), but problems exist at texture-less regions (\eg~walls in 0580 and 0616, floor in 0084) due to the inaccurate scale estimate from sparse reconstructions. We plan to improve these by learning-based sparse or semi-dense reconstruction, \eg~\cite{teed2021droid,teed2022deep}.
\input{table/supp_scannet}

\subsubsection*{Scene-wise statistics on 7-scenes}
\begin{figure*}[h]
    \centering
    \footnotesize
    \begin{tabular}{c@{}c@{}c@{}c}
    \includegraphics[width=.24\textwidth]{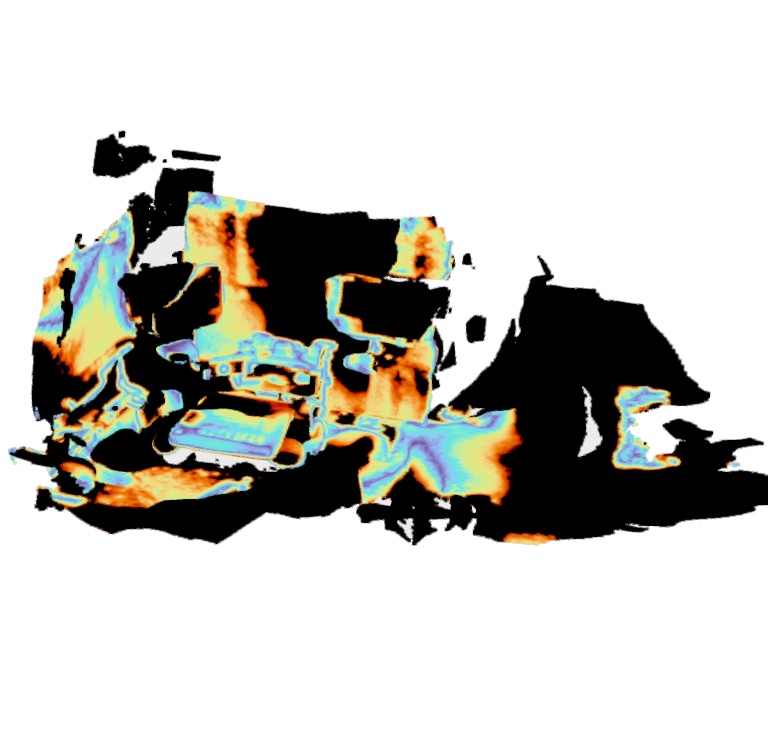} &  
    \includegraphics[width=.24\textwidth]{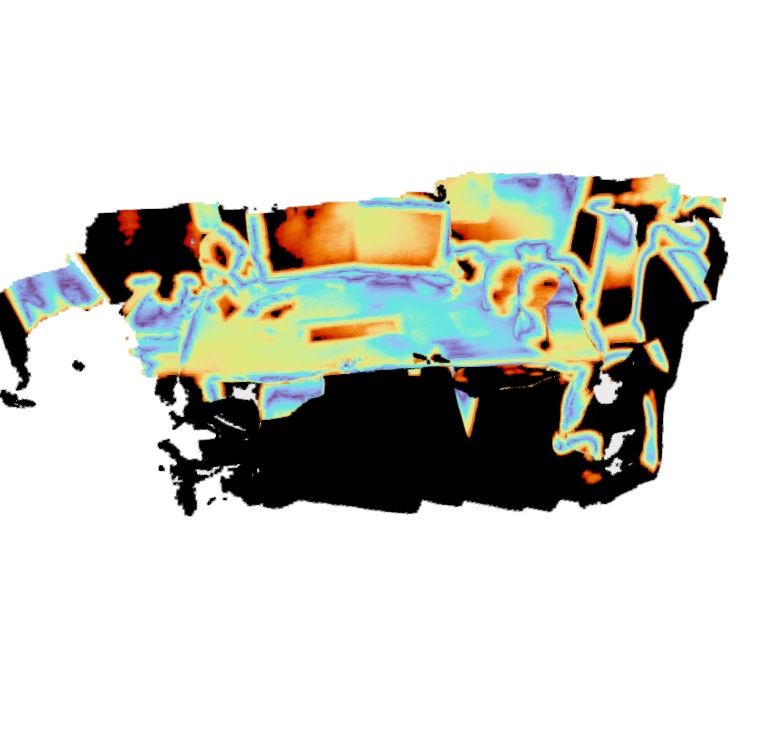} &
    \includegraphics[width=.24\textwidth]{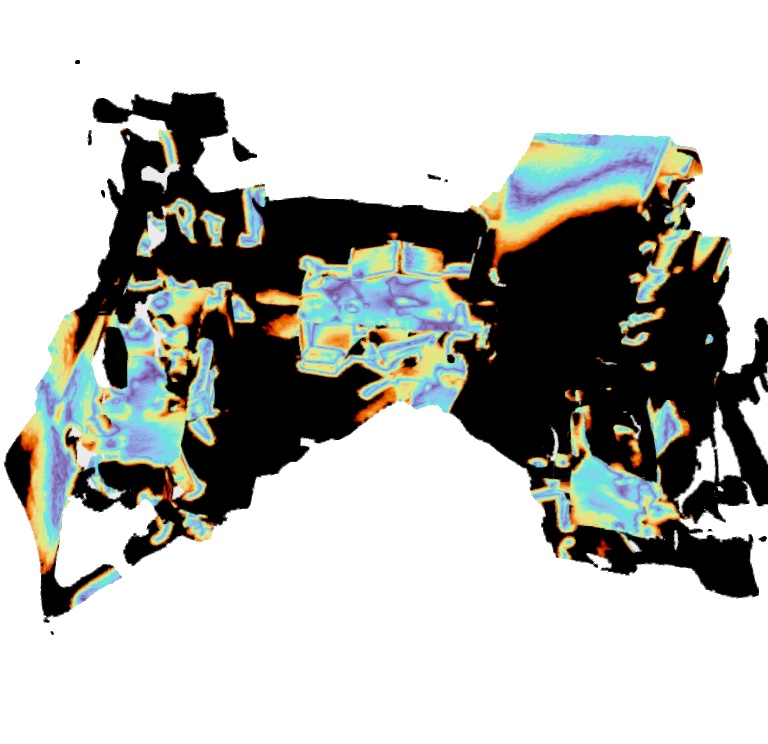} &  
    \includegraphics[width=.24\textwidth]{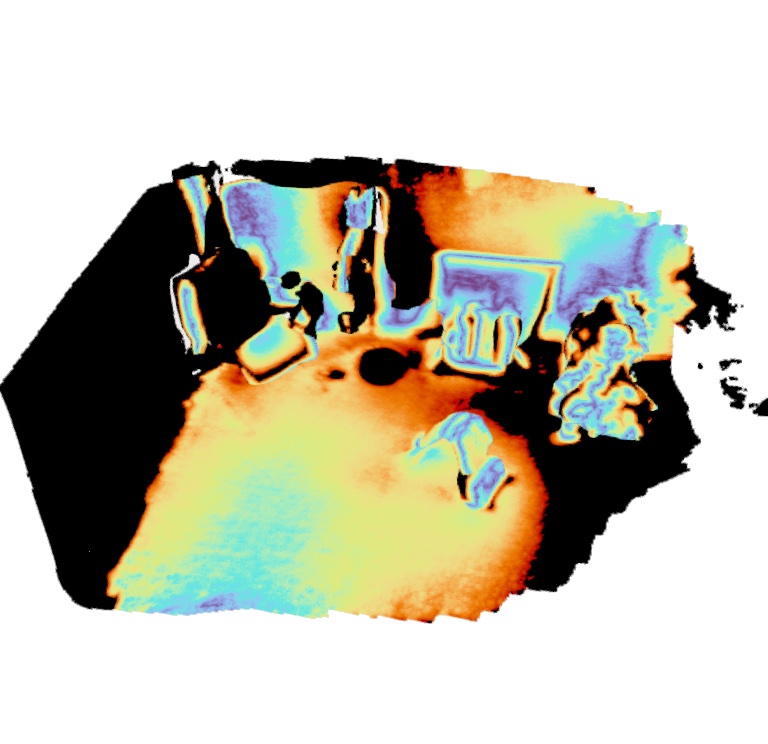} \\    
    \includegraphics[width=.24\textwidth]{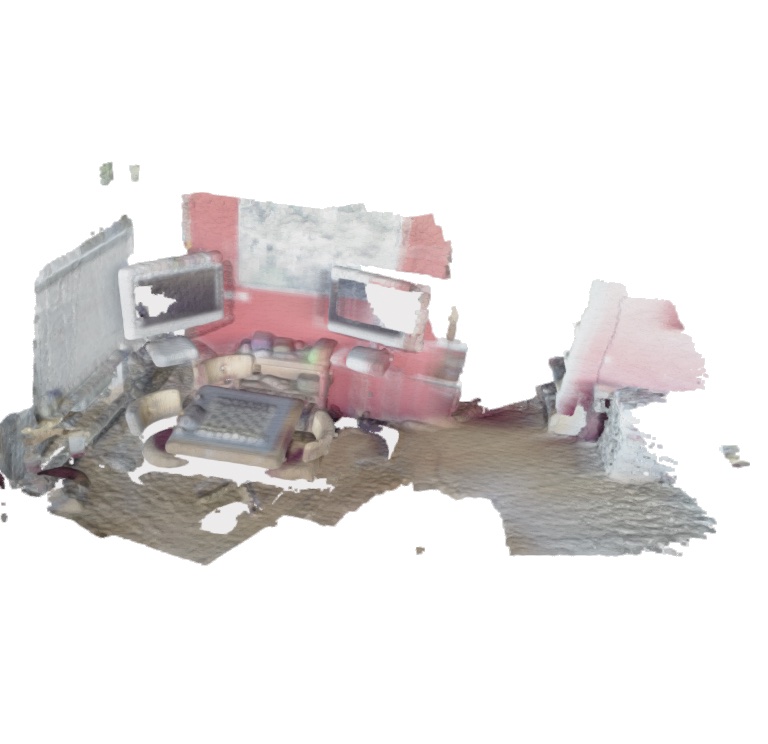} &  
    \includegraphics[width=.24\textwidth]{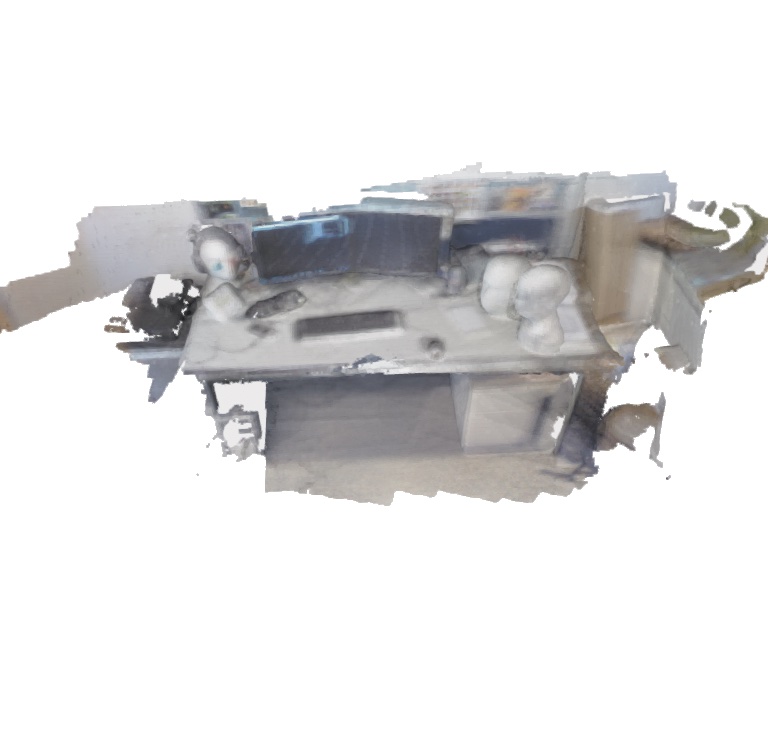} &
    \includegraphics[width=.24\textwidth]{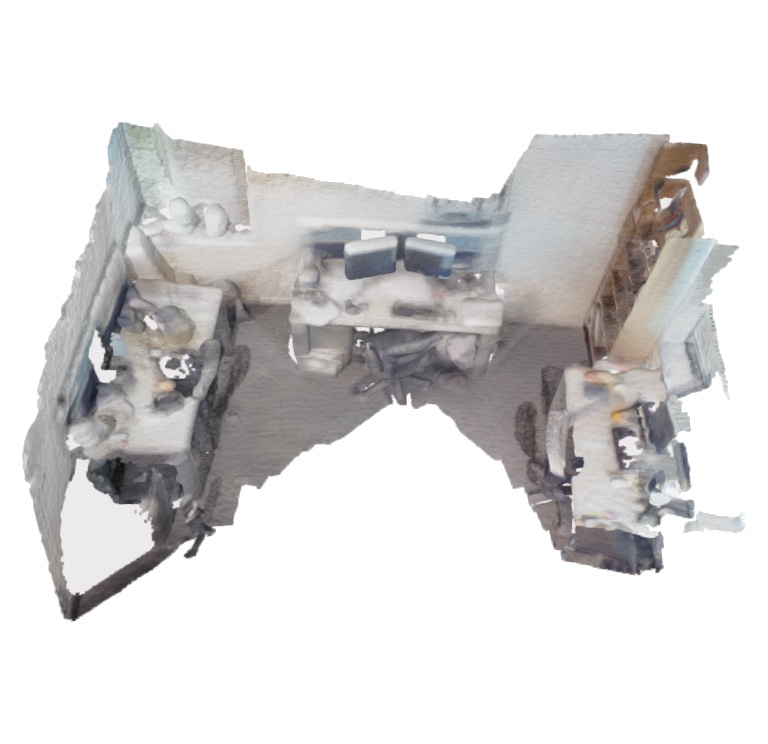} &  
    \includegraphics[width=.24\textwidth]{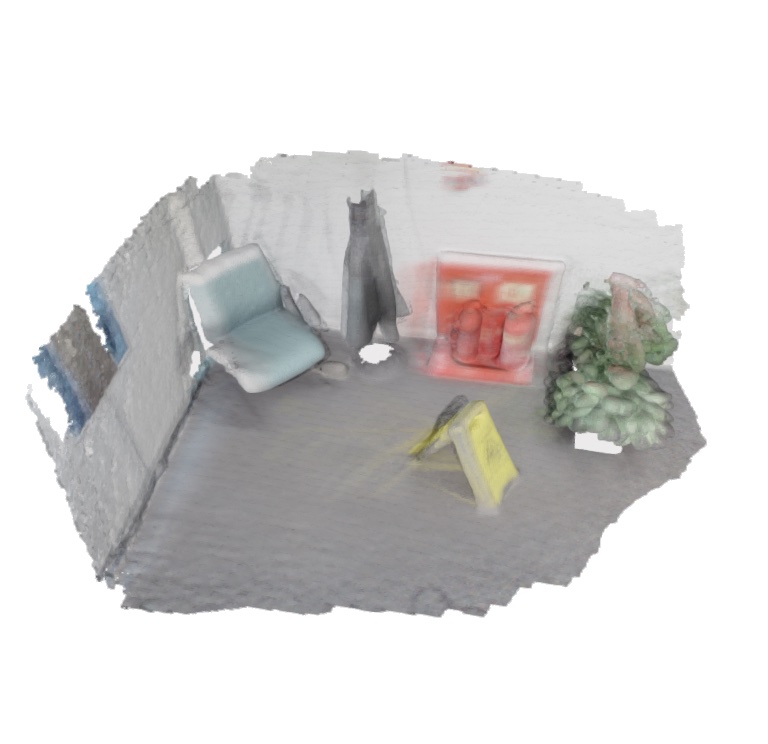} \\        
    chess & heads & office & fire
    \end{tabular}
    \caption{Error heatmap from our reconstruction (first row) to groundtruth (second row) for each scene in 7-Scenes~\cite{glocker2013real}. The colorization is the same as Fig.~\ref{fig:error_map}.}
    \label{fig:error_map_7scenes}
\end{figure*}

The reconstructed mesh and scene-wise statistics are not provided by ManhattanSDF~\cite{guo2022neural} for COLMAP, NeRF, UNISURF, NeuS, VolSDF, and ManhattanSDF. Therefore, we reuse their reported averages as a reference in the main paper.
Here we report scene-wise numbers in Table~\ref{table:recon-accuracy-7scenes} for the state-of-the-art MonoSDF~\cite{yu2022monosdf} and our method. Reconstructions and ground truths are in Fig.~\ref{fig:error_map_7scenes}.

7-scenes have challenging camera motion patterns and complex scenes, thus the overlaps between viewpoints are small, leading to reduced accuracy for all the approaches. Although our approach produces less accurate floor and walls with fewer features, it achieves fine reconstruction of desktop objects in general.

\input{table/supp_7scenes}

%% file: table/supp_scannet.tex
\begin{table*}[htbp]
  \centering
  \caption{Scene-wise quantitative results on ScanNet.}
  \resizebox{.99\textwidth}{!}{
\begin{tabular}{l ccccc | ccccc}
  \toprule
  \multirow{2}{*}{Method} & \multicolumn{5}{c}{0050} & \multicolumn{5}{c}{0084} \\
  \cmidrule(lr){2-11}
  & Acc $\downarrow$ & Comp $\downarrow$ & Prec $\uparrow$ & Recall $\uparrow$  & F-score $\uparrow$  & Acc $\downarrow$ & Comp $\downarrow$ & Prec $\uparrow$  & Recall $\uparrow$  & F-score $\uparrow$  \\
  \midrule
  COLMAP~\cite{schonberger2016structure}    & 0.049 & 0.129 & 0.707 & 0.531 & 0.607 & \two{0.032} & 0.121 & \two{0.807} & 0.577 & 0.673 \\
  NeRF~\cite{mildenhall2021nerf}          & 0.704 & 0.081 & 0.215 & 0.517 & 0.304 & 0.733 & 0.248 & 0.157 & 0.213 & 0.181 \\
  UNISURF~\cite{oechsle2021unisurf}       & 0.432 & 0.087 & 0.309 & 0.482 & 0.376 & 0.594 & 0.242 & 0.218 & 0.339 & 0.266 \\
  NeuS~\cite{wang2021neus}          & 0.091 & 0.103 & 0.528 & 0.455 & 0.489 & 0.231 & 0.365 & 0.159 & 0.090 & 0.115 \\
  VolSDF~\cite{yariv2021volume}        & 0.071 & 0.071 & 0.600 & 0.599 & 0.599 & 0.507 & 0.165 & 0.163 & 0.247 & 0.196 \\
  ManhattanSDF~\cite{guo2022neural}  & 0.032 & 0.050 & 0.849 & 0.755 & 0.800 & \one{0.029} & \one{0.041} & \one{0.822} & \one{0.784} & \one{0.802} \\
  MonoSDF (MLP)~\cite{yu2022monosdf} &  \one{0.025} & {0.054} & {0.865} & {0.713} & {0.781} & {0.036} & 0.048 & {0.700} & 0.646 & {0.672} \\  
  MonoSDF (Grid)~\cite{yu2022monosdf} & {0.027} & \two{0.045} & {0.854} & {0.764} & {0.807} & {0.035} & \two{0.043} & {0.796} & {0.774} & \two{0.785} \\ 
  \midrule
  Ours (Init) & {0.034}  & {0.051} & {0.775} & {0.684}     & {0.727} & 0.047 & {0.048} & {0.705} & {0.725} & {0.715}\\  
  Ours (+Rendering) & \two{0.026}  & \one{0.044} & \two{0.875} & \two{0.780}     & \two{0.825} & 0.038 & {0.046} & {0.762} & {0.748} & {0.755}\\
  Ours (+CRF) & \two{0.026}  & \one{0.044} & \one{0.880} & \one{0.788}     & \one{0.832} & 0.043 & \two{0.043} & {0.750} & \two{0.780} & {0.765}\\  
\midrule \midrule
  \multirow{2}{*}{Method} & \multicolumn{5}{c}{0580} & \multicolumn{5}{c}{0616} \\
  \cmidrule(lr){2-11}
  & Acc $\downarrow$ & Comp $\downarrow$ & Prec $\uparrow$ & Recall $\uparrow$  & F-score $\uparrow$  & Acc $\downarrow$ & Comp $\downarrow$ & Prec $\uparrow$  & Recall $\uparrow$  & F-score $\uparrow$  \\
  \midrule
  COLMAP~\cite{schonberger2016structure}    & 0.169 & 0.300 & 0.204 & 0.112 & 0.145 & 0.045 & 0.406 & 0.689 & 0.230 & 0.344 \\
  NeRF~\cite{mildenhall2021nerf}          & 0.402 & 0.186 & 0.125 & 0.216 & 0.159 & 0.582 & 0.196 & 0.249 & 0.263 & 0.256 \\
  UNISURF~\cite{oechsle2021unisurf}       & 0.392 & 0.192 & 0.131 & 0.188 & 0.155 & 0.571 & 0.148 & 0.237 & 0.300 & 0.265 \\
  NeuS~\cite{wang2021neus}          & 0.206 & 0.275 & 0.167 & 0.114 & 0.135 & 0.137 & 0.140 & 0.330 & 0.289 & 0.308 \\
  VolSDF~\cite{yariv2021volume}        & 0.197 & 0.183 & 0.197 & 0.189 & 0.193 & 0.736 & 0.129 & 0.176 & 0.284 & 0.217\\
  ManhattanSDF~\cite{guo2022neural}  & 0.205 & 0.240 & 0.149 & 0.124 & 0.135 & {0.058} & \two{0.066} & {0.684} & \two{0.513} & \two{0.586} \\
  MonoSDF (MLP)~\cite{yu2022monosdf} &  \one{0.025} & \one{0.040} & \one{0.867} & \one{0.759} & \one{0.809} & \two{0.039} & {0.087} & \two{0.702} & {0.488} & 0.576 \\
  MonoSDF (Grid)~\cite{yu2022monosdf} & \two{0.039} & \two{0.048} & {0.718} & {0.661} & {0.688} & \one{0.033} & \one{0.048} & \one{0.815} & \one{0.646} & \one{0.721} \\ 
  \midrule
  Ours (Init) & {0.076}  & {0.059} & {0.574} & {0.582}     & {0.578} & 0.076 & {0.097} & {0.566} & {0.427} & {0.487}\\  
  Ours (+Rendering) & {0.070}  & {0.080} & \two{0.760} & {0.636}     & {0.692} & 0.046 & {0.070} & {0.699} & {0.504} & \two{0.586}\\
  Ours (+CRF) & {0.046}  & {0.050} & {0.707} & \two{0.682}     & \two{0.694} & 0.057 & {0.080} & {0.659} & {0.504} & {0.571}\\  
  \bottomrule
\end{tabular}
}
\label{table:recon-accuracy-scannet}
\end{table*}

%% file: table/supp_7scenes.tex
\begin{table*}[htbp]
  \centering
  \caption{Scene-wise quantitative results on 7-Scenes.}
  \resizebox{.99\textwidth}{!}{
\begin{tabular}{l ccccc | ccccc}
  \toprule
  \multirow{2}{*}{Method} & \multicolumn{5}{c}{chess} & \multicolumn{5}{c}{heads} \\
  \cmidrule(lr){2-11}
  & Acc $\downarrow$ & Comp $\downarrow$ & Prec $\uparrow$ & Recall $\uparrow$  & F-score $\uparrow$  & Acc $\downarrow$ & Comp $\downarrow$ & Prec $\uparrow$  & Recall $\uparrow$  & F-score $\uparrow$  \\
  \midrule
  MonoSDF (MLP)~\cite{yu2022monosdf} & {0.160} & {0.390} & {0.250} & {0.132} & {0.173} & \one{0.068} & {0.188} & \one{0.586} & {0.353} & {0.440} \\   
  MonoSDF (Grid)~\cite{yu2022monosdf} &  \one{0.113} & {0.143} & {0.324} & {0.267} & {0.293} & {0.133} & {0.099} & {0.305} & 0.327 & {0.315} \\  
  \midrule
  Ours (Init) & {0.164}  & \two{0.108} & {0.278} & {0.350}     & {0.310} & {0.186} & {0.083} & {0.288} & {0.401} & {0.335}\\
  Ours (+Rendering) & \two{0.147}  & {0.111} & \two{0.367} & \two{0.389}     & \two{0.378} & {0.074} & \two{0.062} & {0.543} & \two{0.568} & \two{0.555}\\
  Ours (+CRF) & \two{0.147}  & \one{0.107} & \one{0.368} & \one{0.391}     & \one{0.379} & \two{0.071} & \one{0.057} & \two{0.559} & \one{0.626} & \one{0.591}\\  
\midrule \midrule
  \multirow{2}{*}{Method} & \multicolumn{5}{c}{office} & \multicolumn{5}{c}{fire} \\
  \cmidrule(lr){2-11}
  & Acc $\downarrow$ & Comp $\downarrow$ & Prec $\uparrow$ & Recall $\uparrow$  & F-score $\uparrow$  & Acc $\downarrow$ & Comp $\downarrow$ & Prec $\uparrow$  & Recall $\uparrow$  & F-score $\uparrow$  \\
  \midrule
  MonoSDF (MLP)~\cite{yu2022monosdf} &  \one{0.087} & {0.128} & {0.338} & {0.236} & {0.278} & \two{0.075} & \two{0.064} & \one{0.592} & \two{0.522} & \one{0.555} \\
  MonoSDF (Grid)~\cite{yu2022monosdf} & \two{0.147} & \two{0.077} & \one{0.539} & \two{0.471} & \one{0.503} & \one{0.061} & {0.081} & \two{0.564} & {0.504} & {0.533} \\ 
  \midrule
  Ours (Init) & {0.168}  & \one{0.068} & \two{0.398} & \one{0.483}     & \two{0.436} & 0.087 & \one{0.058} & {0.503} & \one{0.616} & \two{0.554}\\  
  Ours (+Rendering) & {0.180}  & {0.081} & {0.330} & {0.400}     & {0.362} & 0.160 & {0.072} & {0.426} & {0.445} & {0.435}\\
  Ours (+CRF) & {0.164}  & {0.080} & {0.340} & {0.400}     & {0.367} & 0.162 & {0.068} & {0.474} & {0.490} & {0.482}\\  
  \bottomrule
\end{tabular}
}
\label{table:recon-accuracy-7scenes}
\end{table*}